\documentclass{article}

    \usepackage[preprint,nonatbib]{neurips_2024}

\usepackage[utf8]{inputenc} 
\usepackage[T1]{fontenc}    
\usepackage{hyperref}       
\usepackage{url}            
\usepackage{xurl}
\usepackage{booktabs}       
\usepackage{amsfonts}       
\usepackage{nicefrac}       
\usepackage{microtype}      
\usepackage{xcolor}         

\newcommand{\complexity}{LM Evaluation Complexity}
\newcommand{\gpt}{GPT-4}

\usepackage[frozencache=true,cachedir=.]{minted} 

\usepackage{siunitx}

\usepackage{listings}

\author{%
  David S.~Hippocampus\thanks{Use footnote for providing further information
    about author (webpage, alternative address)---\emph{not} for acknowledging
    funding agencies.} \\
  Department of Computer Science\\
  Cranberry-Lemon University\\
  Pittsburgh, PA 15213 \\
  \texttt{hippo@cs.cranberry-lemon.edu} \\
}
\title{Thought of Search: Planning with Language Models \\
Through The Lens of Efficiency}
\author{
Michael Katz\\
IBM Research 
 \And  Harsha Kokel\\
IBM Research 
 \And Kavitha Srinivas\\
 IBM Research 
 \And Shirin Sohrabi \\
 IBM Research 
}

\begin{document}

\maketitle

\begin{abstract}
Among the most important properties of algorithms investigated in computer science are soundness, completeness, and complexity. These properties, however, are rarely analyzed for the vast collection of recently proposed methods for planning with large language models. 
In this work, we alleviate this gap. We analyse these properties of using LLMs for planning and highlight that recent trends abandon both soundness and completeness for the sake of inefficiency. 
We propose 
a significantly 
more efficient approach that can, at the same time, maintain both soundness and completeness. We exemplify on four representative search problems, comparing to the LLM-based solutions from the literature that attempt to solve these problems. We show that by using LLMs to produce the code for the search components we can solve the entire datasets with 100\% accuracy with only a few calls to the LLM. We argue for a responsible use of compute resources;
urging research community to investigate sound and complete LLM-based approaches that uphold efficiency. 
\end{abstract}

\section{Introduction}

Recent work has addressed the issue of planning in Large Language Models (LLMs), spurred by their abilities in natural language tasks.  The approaches vary widely from giving a planning problem to an LLM and asking it to output an entire plan
to asking an LLM to plan step by step, including backtracking.  Roughly, these approaches can be partitioned into two sets. The first exemplifies yet another capability of language models, while the second aims at presenting practical tools for solving planning problems. In the latter case, however, since planning problems are often computationally hard, it is crucial to understand the properties and the complexity of the algorithms proposed. 

The purpose of our work is precisely that. Going over a collection of recent methods for planning with large language models, we analyse the most important properties of the proposed methods such as soundness and completeness. We find all these methods to be neither sound nor complete. 
We further investigate the computational efficiency of these methods in terms of the number of invocations of their most expensive routine -- LLM evaluation.
From a pragmatic perspective, the cost of LLM evaluations is significant, either in terms of GPU resources, or expensive API calls to hosted LLMs.  
We find that abandoning the soundness and completeness does not provide any benefit in computational efficiency, as the proposed methods are prohibitively inefficient, expensive, and most importantly harming the environment. 

We propose an alternative named Thought of Search: thinking {\bf before} searching, an approach that exploits the strengths of LLM, while mitigates the weaknesses of the existing approaches, doing so in an efficient manner. 
We propose using the language models for deriving the symbolically represented search components that allow for performing the search itself without calling LLMs. That way, the search space correctness can be checked before the search is performed, allowing for soundness and completeness (and sometimes optimality) of the search algorithms imply these properties of the overall solution.  Specifically, we focus on the two essential components of any search, successor generator and goal test, using the large language models to obtain their implementation in Python.
We exemplify our proposed approach on four representative search problems, comparing to the LLM-based solutions from the literature that attempt to solve these problems and show these approaches to be prohibitively expensive. We show that by using LLMs, possibly with human feedback, to produce the code for the search components we can solve the entire datasets with 100\% accuracy with only a few calls to the LLM. 
We argue for a responsible use of compute resources;
urging research community to investigate sound and complete LLM-based approaches that uphold efficiency.


\section{Related Work and Beyond: Properties and Complexity}

In this section, we review the related work from the point of view of the soundness and completeness properties of the proposed algorithms, as well as their \emph{\complexity}. We quantify the number of LLM requests required by each algorithm. 
An algorithm is sound if it produces only valid solutions and it is complete if it is guaranteed to produce a solution before terminating successfully.
For consistency, we unify the notation here. In what follows, 
we denote the bound on the number of successors generated per state by $b$, the number of rollouts by $T$, and the length of a rollout/path by $L$.

\paragraph{IO}
The Input-Output (IO) prompting is the straightforward use of LLMs to generate an output for a given input. Some examples that employ IO for planning include querying a pre-trained model \cite{valmeekam-et-al-neurips2023} or fine-tune a model \cite{pallagani-et-al-arxiv2022,SoS}. Here, the model is usually evaluated once, generating an output that may include a sequence of steps, making the complexity of this approach $O(1)$. The method is neither sound nor complete for planning, as it can generate incorrect solutions and not guaranteed to provide a solution if one exists. Further, the methods that fine-tune a model can have a computationally intensive step of data generation.

\paragraph{CoT}
The Chain-of-Thought  approach \cite{wei-et-al-neurips2022} prompts the model with a predefined example of a chain of steps (thoughts) to resolve the question, in an attempt to make the model generate similar chains in its output. Here, still the model is evaluated once, so the complexity is $O(1)$ and 
the approach is still neither sound nor complete for planning, for the same reasons as before.

\paragraph{ReAct}
The ReAct approach \cite{yao-et-al-iclr2023} aims at interleaving  
CoT with acting steps. Each acting step may result in a separate evaluation of the language model. Therefore, the number of evaluations is worst case linear in the number of steps $L$ in the expected output, $O(L)$. This approach is akin to re-planning at each step; or treating LLM as a policy, referenced for each state in the trajectory.  Still, no guarantees of soundness or completeness can be obtained for this approach.

\paragraph{ReWOO}
ReWOO \cite{xu-et-al-arxiv2023} aims at tackling 
the inefficiency in the number of LLM evaluations, 
requesting the first model evaluation to plan all future steps, then each step is executed without model evaluation if possible (and with, if not), finalizing by a model evaluation with the concatenated output of previous states as the input. The best case complexity is therefore $O(1)$, with only two evaluations performed and when external tools are not LLM. The worst case complexity is however still $O(L)$, where $L$ is the plan's length, since each of the $L$ external tool calls can be to an LLM. Same as before, 
no guarantees of soundness or completeness can be obtained.

\paragraph{RAP}
\underline{R}easoning vi\underline{a} \underline{P}lanning (RAP) approach performs a search (e.g., MCTS) using the LLM for expansion (generate successors) and for heuristic or reward prediction of a state 
\cite{hao-et-al-emnlp2023}. Here, the complexity is the worst among the approaches explored so far, being linear in the size of the search space. MCTS is an incomplete algorithm, where the search tree size can be controlled by bounding the number of successors generated per state $b$, number of rollouts $T$,  and their depth $L$. The overall complexity is $O(T\times b\times L)$.
Since the LLM is used for generating successors, it can generate incorrect successors, making the approach also not sound.

\paragraph{ToT}
Tree of Thoughts \cite{yao-et-al-neurips2023} approach is similar to that of RAP, where the “thought generator” is expanding a state by calling an LLM and the state evaluator is calling an LLM to evaluate the generated states. The evaluation can be done per state (one call per state), or across states (one call across all the current generated states). Then, a search is 
performed on the search tree. The worse case overall complexity is $O(N)$, where $N$ is the number of states generated by the search algorithm.
The authors use bounded incomplete versions of the well-known Breadth-First Search (BFS) and Depth-First Search (DFS) algorithms, with the depth bound $L$, branching bound $b$, and beam size $m$, restricting the complexity to $O(b\!\times\!m\!\times\!L)$.
Here as well, the successor generation by LLM eliminates soundness of the otherwise sound algorithms.

\paragraph{GoT}
Graph of Thoughts (GoT) \cite{Besta2023GraphOT} is similar to the tree of thought approach, except that it can connect the thought nodes in the tree above into a graph for more flexible representations through (a) aggregation of thoughts $a$, (b) improvement of a thought to refine it further $i$, (c) generation of a thought based on existing ones $g$.  In GoT, in addition, repeat operation $k$ defines how often a particular operation is repeated.  The complexity is similar to that of the ToT, 
$O(N)$, where $N$ is the number of nodes in the graph generated, in practice bounded by the branching bound $b$ and the depth bound $L$, resulting in the overall complexity of $O(b\!\times\!L)$. Here as well, there are no guarantees of soundness or completeness, as the graph is constructed with the language model.

\paragraph{Reflexion}
Reflection \cite{Reflexion_ShinnCGNY23} is a feedback incorporated approach where in addition to a so-called actor model, which can be either a CoT or a ReAct approach, it has an internal and external feedback component, named an evaluator model and a self-reflection model. The feedback is given within a bounded number of trials and a bounded memory that provides additional context to the agent. The number of evaluations is dependent on which actor model is used plus a constant (two evaluations, one for the evaluator model and one for the self-reflection model) times the number of trials, $T$. So its either $O(T)$ if CoT is used, or $O(L \times T)$ if ReAct is used. While the authors report $100\%$ accuracy given a large enough $T$ in some domains, the approach is not a complete approach. Similar to the other cases, the approach is also not sound.

\paragraph{LATS} Similarly to \cite{hao-et-al-emnlp2023}, Language Agent Tree Search (LATS) \cite{LATS} performs a Monte-Carlo Tree Search (MCTS), using language models as algorithm components. Using our notation of $b$ for the number of successors generated per state, $T$ for the number of rollouts, and $L$ for their depth, the algorithm complexity here as well is $O(T\times b\times L)$. The authors note that in practice, however, their method produce more calls to the LLM than RAP, exploring larger portions of the search space and producing somewhat higher accuracy solutions. As mentioned before, MCTS is already an incomplete algorithm by itself, and LATS limits it further. Since the successors here as well are generated by a language model, the approach is not sound.

\paragraph{AoT}
Algorithm of Thoughts (AoT) \cite{Algo_of_thoughts} combines the CoT and the ToT approach in a following sense: rather than a standard CoT prompting, it incorporates the search strategy (DFS or BFS) within the prompt through the in-context examples. That is they use a standard DFS or BFS to produce the in-context examples.  
The approach is neither sound nor complete, as it relies completely on the language model. Further, it assumes an existence of a search component that can solve the in-context examples. If such component already exists, the only possible reason to use this approach would be if the search component is not efficient, which would imply that the problems are too large to be handled by BFS/DFS (e.g., above $10^{12}$ states). 
But then, 
these search spaces would certainly be beyond any imaginable future capabilities of LLMs. 

While technically AoT makes one call to the LLM per problem instance, the input and output number of tokens grows exponentially with the instance size. Therefore it would be misleading to say that the complexity of this approach is $O(1)$.
From the cost perspective, when charged based on the number of input and output tokens, it might well be more expensive than the other approaches. 
Since we do not have an adequate uniform way of transforming this single call to 
multiple calls of similar size to other approaches, we skip this approach in our analysis.

\paragraph{LLM-Modulo Framework}
The framework suggests to perform planning by a so-called Generate-Test-Critique loop \cite{LLM-Modulo}. The LLM is generating 
candidate plans, which are validated/critiqued by external means, providing feedback to the LLM. These means are assumed to be sound, thus the overall framework is claimed to inherit soundness guarantee. It is not mentioned, however, how to obtain such means of sound validation or critique for cases where no symbolic model already exists. 
Once critiques deemed a plan valid, the algorithm stops. Otherwise, next candidate is generated. There is no guarantee of convergence towards a valid plan and therefore the algorithm is not guaranteed to terminate. This can be fixed by employing a bound on the number of iterations. This, however, does not provide completeness guarantees.


\paragraph{Policy generation with LLMs}
Focusing on problems where no search is required (e.g., polynomial cases) and 
a general policy (or a generalized plan) exists, the authors propose querying LLMs to provide a policy implementation in python directly \cite{silver-et-al-aaai2024}. The policy is then checked on a few held out problem instances from the domain, providing a feedback on problems that are not solved by the policy. The number of calls to the LLM is therefore $O(1)$ \textbf{per domain}, and the result can be amortised among all the problem instances of that domain. 
This approach, while tackling the simpler case of problems where no search is needed, has served as an inspiration for our proposed approach.

\section{Proposed Approach}
In this work, we argue for an efficient and responsible use of compute resources. 
First and foremost, this means using the computationally expensive large language models efficiently and appropriately. 
In the case of solving search problems, as in the focus of this work, we argue against using an LLM at each expansion and evaluation.
Such an implementation of search algorithms is inefficient, but equally importantly, it also sacrifices important properties that search algorithms possess, such as soundness and completeness.
While completeness is often sacrificed knowingly, in order to limit the number of possible calls to the language model, soundness can be lost unintentionally. 
If the state successors are generated by a large language model, there is no guarantee that the produced successors are {\em valid}. An additional mechanism that validates the produced output would be required to render the algorithm sound.
Such a mechanism would be symbolic by nature, since it must guarantee correctness. 
However, if such mechanism exists, it may be possible to use that mechanism to produce successors, without the need for performing the calls to a large language model at each evaluation. The large language models would be a prime candidate for producing such a mechanism.
%
%
The mechanism can be a symbolic planning model, e.g., 
Planning Domain Definition Language (PDDL), 
from which all the search components can be computed, as is done by classical planners that perform heuristic search \cite{helmert-jair2006}. 
Alternatively, large language models can also directly produce the search components code: a successor function, a goal test, and even a heuristic function 
\cite{hart-et-al-ieeessc1968} or reward. While the former approach has been explored in the literature \cite{guan-et-al-neurips2023,oswald-et-al-icaps2024}, here we 
focus on the latter. We propose to use large language models
for obtaining a Python implementation of two critical search components, {\em successor functions} and {\em goal test}. We query the language model for each component separately, using a textual description of the problem at hand. We assume the process to be iterative and user-guided, by a person capable of validating the code obtained, leaving automation of the process for future work. 
It is worth noting that if the expensive evaluation of large language models is not performed during search, there is no need to artificially restrict the algorithms to their incomplete variants. Still, the code must be validated for soundness, ensuring that all produced successors are correct, as well as completeness, ensuring that all possible immediate successors are produced. 

\section{Experiments}
We exemplify the proposed approach with a variety of popular search problems, demonstrating the use of large language models to obtain the search components code for a representative set of the same search problems that the vast recent work on planning using LLMs used. Our intention is to exemplify the efficiency and accuracy gap created by the unsupervised use of large language models at every search step. For each search problem, we count the number of interactions with the language model to obtain valid\footnote{Valid from the perspective of a software developer who understands the problem solved.} successor function and goal test implementations. We then run a standard implementation of either a BFS or a DFS algorithm with these two functions on a set of instances and report the accuracy and the total time to solve the instances. 
We repeat the experiment 5 times, obtaining 5 implementation variants and compare the efficiency of these variants in terms of total time. Note that if the implementation of successor function and goal test obtained from the large language model is correct, the accuracy is guaranteed to be 100\% (if sufficient resources are given to the search). The opposite is not necessarily true, as the set of tested instances might not include some corner cases that make the code perform incorrectly. However, we see that as one of the strengths of our approach, as such cases can be used for testing and debugging the obtained code. Under the assumption of correctness of the successor function and goal test implementations, our approach inherits the soundness and completeness guarantees of the search algorithms used \cite{cormen-et-al-1990}. In all our experiments we use the \gpt~ model \cite{openai2024gpt4}, in a chat mode. The code obtained from \gpt\ in all 5 correspondences is provided in the appendix. The search with the obtained code was run locally, on a machine with 2.3 GHz 8-Core Intel Core i9 CPU, no GPUs were used.

\begin{listing}[!t]
  \scriptsize
  {\em User}:
  The 24 Game is a mathematical card game in which the objective is to find a way to manipulate four integers so that the end result is 24. The game is played with a list of four numbers, and the player must use all four numbers exactly once, using any combination of addition, subtraction, multiplication, or division, to arrive at the number 24. If a state is defined by a list of numbers (4 or less), how are successor states defined? Provide only the python code that returns a list of successor states for a state. 
  
  \vspace{0.2cm}
  
  {\em ChatGPT}:
  Below is the Python code that defines how to generate the successor states for a given state in the 24 Game. A state in the 24 Game is a list of numbers. The successor states are generated by taking two numbers from the state, applying an operation (addition, subtraction, multiplication, or division if the divisor isn't zero), and then forming a new list with the result and the remaining unused numbers.
  
  \begin{minted}{python}
  import itertools
  
  def generate_successor_states(state):
      successors = []
      n = len(state)
      for (i, j) in itertools.combinations(range(n), 2):
          a, b = state[i], state[j]
          remaining = [state[k] for k in range(n) if k != i and k != j]
          
          # Generate all possible new states
          successors.append(remaining + [a + b])
          successors.append(remaining + [a * b])
          if a > b:
              successors.append(remaining + [a - b])
          else:
              successors.append(remaining + [b - a])
          if b != 0:
              successors.append(remaining + [a / b])
          if a != 0:
              successors.append(remaining + [b / a])
      
      return successors
  \end{minted}
      
  \vspace{0.1cm}
  This function takes a state (a list of numbers), iterates over all pairs of numbers, applies the operations to these pairs, and then forms new lists with the result of the operation and the other numbers that were not used in the operation. This ensures that each successor state is a valid step towards finding a solution to reach 24, using all four numbers exactly once along with any of the allowed operations.
  
  \vspace{0.2cm}
  
  {\em User}:
  Provide only the python code that tests whether a state is a goal state.
  
  \vspace{0.2cm}
  
  {\em ChatGPT}:
  Here's the Python code that tests whether a given state is a goal state in the 24 Game. A goal state is achieved if the list contains exactly one number, and that number is 24.
  
  \begin{minted}{python}
  def is_goal_state(state):
      return len(state) == 1 and state[0] == 24
  \end{minted}
  
  This function checks if the length of the state list is 1 and whether the sole element in the list is the number 24.
  \caption{\textbf{24 game} example correspondence to obtain the successor function and the goal test.}
  \label{24game}
  \end{listing}
  
\paragraph{24 game} The 24 game \cite{yao-et-al-neurips2023} takes 4 integers as an input that can be manipulated through the four most common arithmetic operations: addition, subtraction, multiplication, and division. The goal of the game is to produce a formula (also representable by a sequence of operations) that evaluates to 24, if one exists. As mentioned before, \cite{yao-et-al-neurips2023} achieve that with an incomplete algorithm that 
performs $\approx100$ calls to the LLM, according to their report.
The authors report a success rate of $75\%$ on a subset of 100 medium difficulty problems out of a collection of 1362 problems \cite{yao-et-al-neurips2023}.

As mentioned before, we propose instead to use the LLM to produce the code for two functions: {\em successor function} and {\em goal test}. We leave the heuristic function to future work, noting that even without it, the search space of the 24 game is rather small, up to $1 + 36 + 648 + 3888 = 4573$ states. 
Listing \ref{24game} describes an example correspondence with the LLM. 
We observe that the model required 1.2 interactions on average for the successor function and 1 interaction for the goal test. We see that, with a minor help from the user, the model is able to produce correct successor and goal test functions for this problem. 
We further tested all 5 successor functions and goal test within a simple BFS implementation on the entire suite of 1362 instances from ToT \cite{yao-et-al-neurips2023}. 
For each of the 5 tested successor functions, BFS found a solution in the 1361 cases where a solution exists and report that no solution exists in the only one unsolvable case, a 100\% success rate.
%
The total time to solve all 1362 problems varies over these 5 cases from 
1.92s to 6.83s
in our naive BFS implementation, hinting that some successor functions can be more efficient than other. This is comparable to a \textbf{single} LLM evaluation time; which is $\approx7s$ for \gpt~ Chat~\cite{openaiPerformanceAnalysis}. 
Note that the generated successor functions are generic enough to be able to solve the generalized version of the 24game, Countdown \cite{SoS}, with only minimal adaptation to the goal test.

\paragraph{Mini crosswords} The mini crosswords \cite{yao-et-al-neurips2023} is a 5x5 crosswords dataset that includes 20 games, where the input describes the 5 horizontal and 5 vertical clues and the output is the full 25 letters board. 
%
%
We used \gpt~ to produce the two functions: {\em successor function} and the {\em goal test} and repeated the experiment 5 times.
The correspondence with the model can be found in the appendix.
We observe that the model required 2.4 interactions on average to produce a valid successor function, and 1.4 interactions on average to  produce the goal test, with errors primarily related to not handling the corner cases of incorrect input. 
In all cases, after providing the exact error to the model, it added safeguards which fixed the issues. 
We tested the obtained functions within a standard implementation of a DFS with a closed list on the entire suite of 20 games \cite{yao-et-al-neurips2023}. As our focus in this work is on the search aspects, we assume that each clue in each crossword has 10 variants of possible answers, including the correct one.
All 20 games were solved by all five generated pairs of implementation of successor function and goal test (100\% accuracy), with a total time for all 20 games varying from 5.5s to 346s, yet again signifying the importance of efficient implementation of the successor function.

To compare to the ToT approach, 
with the bound of $100$ on state expansions, it calls the LLM $\approx200$ times 
in the worst case 
for each of the 20 games. To be fair to the ToT approach, it does not assume the availability of possible answers, deriving the candidates, at each step, with the help of the language model. This, however is a mixed blessing, since the language model does not provide a guarantee of including the correct answer among produced variants.




\paragraph{BlocksWorld} Probably the most famous planning domain is BlocksWorld, where the blocks can be picked up from the table, put down on the table, unstacked from other blocks or stacked on other blocks to transform the initial configuration to a given goal configuration. The domain has a known planning model, described in PDDL and it is one of the tasks considered by the reasoning-as-planning approach \cite{hao-et-al-emnlp2023}. 
As in the previous cases, we use the \gpt~ model to obtain the successor function and the goal test implementation in Python. We use the same textual description of the domain as \cite{hao-et-al-emnlp2023}. 
A correct successor function and goal test are obtained after 2.8 and 1 iterations on average, respectively. 

  \begin{listing}[!t]
    \scriptsize
    {\em User}:
    Logical deduction is performed by applying rules of the form if X then Y to our current knowledge about a world, extending our knowledge with each rule application. A rule if X then Y is applicable if we know X. If a state is defined as our current knowledge of the world, how are successor states defined, given a collection of rules? Provide only the Python implementation of successor states function.
    
    \vspace*{0.1cm}
    {\em ChatGPT}:
    %
    %
    %
    %
    %
    %
    \begin{minted}{python}
    def successor_states(current_state, rules):
        """
        Generate all successor states from the current state by applying applicable rules.
        
        Args:
        current_state (set): A set of facts known in the current state.
        rules (list of tuples): A list where each tuple represents a rule in the form (conditions, conclusion),
                                where `conditions` can be a single fact or a set of facts.
        
        Returns:
        list of sets: A list of all possible successor states.
        """
        new_states = []
        for conditions, conclusion in rules:
            # Check if the rule's conditions are met in the current state
            if isinstance(conditions, set):
                if conditions.issubset(current_state):
                    # Apply the rule: create a new state with the conclusion added
                    new_state = current_state.union({conclusion})
                    new_states.append(new_state)
            else:
                if conditions in current_state:
                    new_state = current_state.union({conclusion})
                    new_states.append(new_state)
    
        return new_states    
    \end{minted}
    
    
    {\em User}:
    Provide only the python code that tests, given a target fact, whether a state is a goal state.
    
    \vspace*{0.1cm}
    
    {\em ChatGPT}:
    
    \begin{minted}{python}
    def is_goal_state(state, target_fact):
        """
        Check if the state contains the target fact, indicating it is a goal state.
        
        Args:
        state (set): The current state represented as a set of facts.
        target_fact (str): The fact that defines the goal state.
        
        Returns:
        bool: True if the target fact is in the state, False otherwise.
        """
        return target_fact in state
    \end{minted}
    
    
    \caption{\textbf{PrOntoQA} example successor states function and goal test (non-code text removed).}
    \label{prontoqa}
    \end{listing}
    
The mistakes \gpt~ makes when producing the code repeat from one experiment to another, and are often easy to fix inline, without additional iterations (e.g., using shallow copy when deep copy is needed). In our experiments, however, we did count these as additional iterations.
In order to evaluate the obtained functions, we used them within a standard implementation of a BFS and experimented with the collection of 502 instances from \cite{hao-et-al-emnlp2023}. All 502 tasks were solved by all five generated pairs of successor function and goal test (100\% accuracy), with the total time for all 502 tasks varying from 0.56s to 9.7s. The more time efficient approaches represented a state as a set of strings representing boolean variables (or logical predicates), while the less efficient representation used dictionaries of lists of objects or pairs of objects with predicates as keys. The simpler state representation also resulted in a more efficient duplicate detection in our rather naive implementation.

Note that the accumulated number of expanded\footnote{A state is expanded if its successors are generated.} and generated states in the most efficient case (with duplicate detection) was 50143 and 129408, respectively. 
If we needed to call the \gpt~ model on each expansion and generation, by the most conservative estimation, it would take approximately 14 days and cost over $\$1000$, while not guaranteeing correctness of the outcome. Note that the instances considered are quite small - they have only 4 to 5 blocks. Larger BlocksWorld instances would require significantly more resources. This is true for the proposed approach as well, and larger instances mean larger state spaces, becoming too large for an uninformed search such as BFS. In such cases, a heuristic function may be of help, either, similarly, implemented in Python or automatically derived from a PDDL representation, which in turn may be obtained with the help of a large language model \cite{oswald-et-al-icaps2024}.  
It is also worth noting that BFS not only guarantees that the obtained solution is correct, but also guarantees that it is optimal, while LLMs can never provide such a guarantee - checking whether a given solution is optimal is as hard as finding an optimal solution.

\paragraph{PrOntoQA} Logical reasoning can be viewed as a search problem of finding a sequence of logical rules that when applied to the known facts, derive or disprove the target hypothesis. Previous work applies Monte-Carlo Tree Search (MCTS) with successor function and rewards obtained by calling an LLM, to examples from the PrOntoQA dataset to derive the answer but also the proof, a sequence of reasoning steps \cite{hao-et-al-emnlp2023}. The authors report performing 20 iterations for MCTS and 20 samples for self-consistency, resulting in 94.2\% correct answer rate and 78.8\% proof accuracy. 

Similarly to the previous cases, we have generated the successor function and the goal test with the help of \gpt~ model and obtained the answer and the proof by running BFS at most twice per question: once trying to prove the positive hypothesis and, if not successful, once more trying to prove the negation of the hypothesis. The search, if successful, returns a path from the initial state to a found goal state, which corresponds precisely to the sequence of applications of the reasoning rules -- the proof. We performed the experiment 5 times, with the language model being able to produce a correct successor function and goal test after 1.6 and 1 iterations on average, respectively. We tested the 5 obtained pairs of functions on the entire collection of 4000 questions generated by \cite{hao-et-al-emnlp2023}. All 4000 questions were answered correctly, with all generated proofs guaranteed to be valid, resulting in 100\% accuracy for both, with the total time for all 4000 questions varying between 2.16s and 2.53s.  Listing \ref{prontoqa} shows an example correspondence with the model, with non-code text removed for brevity. 


\section{Discussion}

It is hard to overstate the importance of the ability to solve search problems and it is natural to solve these problems by exploring some portion of their state space. All the methods we discuss here do that in one way or another. The difference between those approaches is in how big is the portion of the explored state space and what is the cost of exploring that portion. As all the discussed approaches are greedy, unsound, and incomplete, the accuracy of the solutions they provide can only be evaluated experimentally for particular problem and a dataset, and the results do not reflect on other problems or even datasets within the same problem. 

In this section we perform a thought experiment. We ignore the accuracy of the approaches and only compare their estimated cost and the portion of the search space they have the potential to explore. 
For each approach we estimate two values. First, we estimate the number of calls to a language model performed if the approach was run on the four datasets from the previous section, as a proxy for the cost of using the approach.  Second, we estimate the number of states explored during search, as a proxy for a portion of the state space size. The actual state space size for these datasets can be calculated precisely, and we provide these values. Note that the number of states explored is an over-estimation for the portion of the state space explored, since the same state can be explored several times. This is true for sound and complete search algorithms as well. These algorithms often employ a duplicate detection mechanism to prevent re-evaluating and re-expanding such states, when such re-expansions are not needed to guarantee algorithm properties (e.g., optimality when admissible but inconsistent heuristics are used).  
We also provide the actual number of calls to the language model and the number of states explored using our proposed approach. The actual number of calls for our proposed approach is the number of calls that was sufficient for obtaining soundness and completeness (and sometimes optimality). The number of states explored indicated an upper bound on the portion of the states space necessary for exploration.

\paragraph{Datasets size}
We use the same four representative search problem datasets from previous section. The size of each dataset is denoted by $D$. 
For the 24 game, $D$ is $1362$, for Crossword it is $20$, for BlocksWorld it is $502$, while for PrOntoQA it is $4000$. We used the datasets provided by the approaches we compare to, but these numbers could also have been significantly larger.
For these datasets, we have also computed the size of the search problems in terms of the summed number of states over the instances in the dataset. For the 24 game, the number of states per instance is $4,573$ and therefore the summed number of states is $6,228,426$. For Crossword, the number of states is $\sum_{i=0}^{10} {n \choose i} 10^i=(10 + 1)^{10}$ per puzzle, and therefore the summed number of states is $518,748,492,020$. For BlocksWorld, our dataset includes $447$ instances with $4$ blocks and $55$ instances with $5$ blocks. The instances with $4$ blocks have $125$ states each and the instances with $5$ blocks have $866$ states each, giving us the total of $103,505$ states. Finally, for PrOntoQA, the number of states per task varies between $12$ and $54$, with the total summed number being $97,608$.

\paragraph{Uniform restriction of the search space}
Each approach sets limitations to restrict the number calls to the language model, which makes it difficult to compare. In this thought experiment, we choose to use only the bounds on the breadth/depth of the search and the number of trials, ignoring the absolute constant bounds on the number of explored states used by some approaches.
For a fair comparison across the approaches, we will use the same parameter value for all, even if the original work used a different value. We aimed at smaller values among used across the different approaches for each parameter. This results in under-approximation of the number of calls in most cases, and possibly would have resulted in a lower than reported by the approaches accuracy. In this experiment, however, we ignore the accuracy of the approaches.
We use the branching bound $b=5$, the number of trials $T=10$, and the beam size $m=5$
.
The only parameter we set according to the dataset is $L$, the bound on the sequence length. This is due to the fact that if $L$ is not sufficiently large, the algorithms will not be able to find a solution.
In the 24 game $L=3$ is sufficient, while in Mini crosswords it must be at least $10$. In both BlocksWorld and PrOntoQA the length of a plan/proof varies, with the maximal length over the instances in the datasets in BlocksWorld being $16$ and in PrOntoQA being $6$. 

\begin{table}[!t]
  \setlength\tabcolsep{5.0pt} 
  \def\arraystretch{1.08}
  \centering
  \begin{tabular}{|l|l|rr|rr|rr|rr|}
  \multicolumn{2}{c|}{} & \multicolumn{2}{c|}{24Game} & \multicolumn{2}{c|}{Crossword}& \multicolumn{2}{c|}{BlocksWorld}& \multicolumn{2}{c|}{PrOntoQA}\\
  Approach	  &   Complexity & States& Calls &   States& Calls  & States& Calls &  States& Calls \\
   \hline
  IO & O(D) & 0.02\% & 1362 & 4e-9\% & 20 &0.5\%& 502 &4\%& 4000 \\ 
  CoT & O(D) & 0.02\% & 1362 & 4e-9\% & 20 &0.5\%& 502 &4\%& 4000 \\
  ReAct & O(LD)& 0.07\% & 4086 &4e-8\%& 200 &7.8\%& 8032 &24.6\%& 24K \\
  ReWOO & O(LD)& 0.07\% & 4086 &4e-8\%& 200 &7.8\%& 8032 &24.6\%& 24K \\
  RAP & O(TbLD)& 3.3\% & 245K &2e-6\%& 12K &388\%& 482K & 1229\% & 1.44M \\
  ToT & O(bmLD)& 1.6\% & 102K &1e-6\%& 5K &194\%& 201K & 615\% & 600K\\
  GoT & O(bLD) & 0.3\% & 20K &2e-7\%& 1K &39\%& 40K & 122\% & 120K\\
  Reflection & O(LTD) & 0.7\% & 68K  &4e-7\%& 2.4K &77.6\%& 90K& 245\% &320K\\
  LATS & O(TbLD) & 3.3\% & 286K  &2e-6\%& 14K &388\%& 562K & 1229\% & 1.68M\\
   \hline
  ToS (ours) & O(1)  & 27.0\% & {\bf 2.2} & 3e-4\% & {\bf 3.8} & 125\% & {\bf 3.8} & 175\%& {\bf 2.6}
  \end{tabular}
  \vspace{0.2cm}
  \caption{\label{complexity} Comparison of various approaches on evaluated tasks. The number of tasks is $D$, the length bound of the search/rollout/plan is $L$, number of rollouts is $T$, beam size is $m$, and branching bound is $b$. The summed number of states over the $D$ tasks and the projected number of LLM evaluations are given per approach and problem. 
  }
  \end{table}

\paragraph{Analysis}
The complexity analysis performed in the previous section does not tell the whole story. It hides the constant multiplier, which in some cases could be significant. We also need an additional information about the number of states traversed. Let us take another look at the explored approaches.  

Both IO and CoT only call a language model once per instance. Assuming that the output is a plan, it traverses $L$ states.
ReAct and ReWOO (worst case) make exactly $L$ calls to the LLM, exploring $L$ states. 
Both RAP and LATS perform MCTS, but the actual number of calls to the LLM varies. RAP performs $LT + bLT$ calls (see Algorithm 1 in \cite{hao-et-al-emnlp2023}), while LATS performs $2LT + bLT$ calls (see Algorithm 1 in \cite{LATS}). The number of states explored in both cases is $bLT$.
Reflection with ReAct performs $(2+L)T$ calls, exploring $LT$ states.
ToT introduced their own variants of well-known BFS and DFS algorithms with bounds on the branching factor, search depth, but also a limit on the open list size (beam size). While their performance can vary greatly on individual tasks, with DFS typically being greedier and therefore faster, in the worst case they both call the LLM $bmL$ times and explore $bmL$ states. The GoT approach calls the LLM $bL$ times, exploring $bL$ states. 

Table \ref{complexity} shows the estimated numbers according to the parameters values and the analysis above. 
For comparison, the last row of the table depicts our proposed approach. 
\%States columns denotes the portion of the state space explored. 
For our approach, this is the actual value from our experiments. Recall, in the case of PrOntoQA the BFS search is performed once or twice, until the hypothesis or its opposite is proven. 
For the other approaches, this is an estimate, under the assumptions that the search does not explore the same states multiple times. 
Clearly, this assumption does not hold in practice, with the methods exploring the same state many times. This is also true for a sound and complete search, albeit to a lower extent, it can generate the same state multiple times. An important property of a sound and complete search is that it generates states {\em systematically}, and expands them at most once.
The number of times the same state is generated tends to be higher in MCTS than in BFS/DFS. Non-systematic successor generation
is also a major contributor to that inefficiency. Having in mind that these numbers are very crude over-approximations, we observe that the investigated approaches explore only a very small portion of the search space. This is one of the major sources of their low accuracy -- if the solution is not completely included in that explored portion, then there is no way for that solution to be found. In 24 game and crossword, the portion explored tends to be very low, one or two orders of magnitude smaller compared to our method. In BlocksWorld it is comparable to our method, and in PrOntoQA it is often much higher, order of magnitude larger than the size of the state space, indicating that the same states are explored over and over again.

Looking at the overall number of calls, some methods are more expensive than other. Assuming an average of 500 tokens per input\footnote{We do not consider in our analysis the length of the input, which is a crucial factor of the evaluation cost.} and 50 tokens per output, according to the current pricing of \gpt-turbo (the cheaper of \gpt~ models), the overall cost varies from $\$40$ for IO/CoT and $\$200$ for ReAct/ReWOO, to $\$14,000$ for RAP and $\$16,000$ for LATS.

\section{Conclusions and Future Work}

The current trends in planning with large language models focus on performing a search when the search components are realized through the large language models. 
We analyze the existing approaches and show them to be unsound, incomplete, and quite expensive. We propose to use the large language models to instead generate a code for these search components, to be reused throughout the entire dataset. 
%
The significant differences in the use of computational resources and the performance measures clearly demonstrate that LLMs can be effectively employed for planning purposes without compromising on soundness, completeness, efficiency or accuracy. 

For future work, we would like to explore the use of large language models to obtain the code for search guidance or search pruning techniques. But more importantly, we would like to relax the need for human feedback in coming up with valid implementations of the search components. This could be done in ways similar to the way a policy is obtained in generalized planning \cite{silver-et-al-aaai2024}.


\bibliographystyle{plain}

\clearpage 

\appendix

\section{Search implementation}

The search components {\em successor\_states} and {\em goal\_test} are plugged into the search algorithms. In this work we used BFS and DFS algorithms, implemented as follows.

\begin{minted}{python}
def _str(state):
    if isinstance(state, dict):
        return str(state)
    return " ".join(sorted(list([str(s) for s in state])))


def reconstruct_plan(s, Closed):
    plan = []
    current = s
    while current is not None:
        plan.append(current)
        c = _str(current)
        current = Closed[c]
    return plan[::-1]
    

def bfs(state, successor_states, is_goal):
    expanded = 0
    generated = 0
    s = state
    Q = [tuple((s, None))]
    Closed = dict()
    while len(Q) > 0:
        # Get the top from the queue
        s, parent = Q[0][0], Q[0][1]
        del Q[0]
        c = _str(s)
        if c in Closed:
            continue
        Closed[c] = parent
        if is_goal(s):
            return reconstruct_plan(s, Closed), expanded, generated
        expanded += 1
        for t in successor_states(s):
            Q.append(tuple((t,s)))
            generated += 1
    return None, expanded, generated  


def dfs(state, successor_states, is_goal):
    expanded = 0
    generated = 0
    s = state
    Q = [tuple((s, None))]
    Closed = dict()
    while len(Q) > 0:
        # Get the top from the queue
        s, parent = Q[-1][0], Q[-1][1]
        del Q[-1]
        c = str(s)
        if c in Closed:
            continue
        Closed[c] = parent
        if is_goal(s):
            return reconstruct_plan(s, Closed), expanded, generated
        expanded += 1
        for t in successor_states(s):
            Q.append(tuple((t,s)))
            generated += 1
    return None, expanded, generated
\end{minted}

\section{Model correspondences}

\subsection{24 game}

The prompt for the successor function is as follows.\\
{\em  The 24 Game is a mathematical card game in which the objective is to find a way 
    to manipulate four integers so that the end result is 24. The game is played 
    with a list of four numbers, and the player must use all four numbers exactly 
    once, using any combination of addition, subtraction, multiplication, or 
    division, to arrive at the number 24. If a state is defined by a list of numbers
    (4 or less), how are successor states defined? Provide only the python code that 
    returns a list of successor states for a state. }

The prompt for the goal test is as follows.\\
{\em Provide only the python code that tests whether a state is a goal state.
}

\subsubsection{Experiment Run 1.}
{\em User}:
The 24 Game is a mathematical card game in which the objective is to find a way to manipulate four integers so that the end result is 24. The game is played with a list of four numbers, and the player must use all four numbers exactly once, using any combination of addition, subtraction, multiplication, or division, to arrive at the number 24. If a state is defined by a list of numbers (4 or less), how are successor states defined? Provide only the python code that returns a list of successor states for a state. 

\vspace{0.2cm}

{\em ChatGPT}:

\begin{minted}{python}
import itertools

def successor_states(state):
    if len(state) == 1:
        return []

    successors = []
    for i, j in itertools.combinations(range(len(state)), 2):
        remaining = [state[k] for k in range(len(state)) if k not in (i, j)]

        pairs = [
            state[i] + state[j],
            state[i] - state[j],
            state[j] - state[i],
            state[i] * state[j],
        ]

        if state[j] != 0:
            pairs.append(state[i] / state[j])

        if state[i] != 0:
            pairs.append(state[j] / state[i])

        for result in pairs:
            successors.append(remaining + [result])

    return successors
\end{minted}
    
\vspace{0.2cm}

{\em User}:
Provide only the python code that tests whether a state is a goal state.

\vspace{0.2cm}

{\em ChatGPT}:

\begin{minted}{python}
def is_goal_state(state):
    return len(state) == 1 and abs(state[0] - 24) < 1e-9    
\end{minted}


\subsubsection{Experiment Run 2.}
{\em User}:
The 24 Game is a mathematical card game in which the objective is to find a way to manipulate four integers so that the end result is 24. The game is played with a list of four numbers, and the player must use all four numbers exactly once, using any combination of addition, subtraction, multiplication, or division, to arrive at the number 24. If a state is defined by a list of numbers (4 or less), how are successor states defined? Provide only the python code that returns a list of successor states for a state. 

\vspace{0.2cm}

{\em ChatGPT}:

\begin{minted}{python}

from itertools import permutations, product

def apply_op(a, b, op):
    if op == '+':
        return a + b
    elif op == '-':
        return a - b
    elif op == '*':
        return a * b
    elif op == '/':
        if b != 0:
            return a / b
        else:
            return None

def generate_successor_states(state):
    if len(state) == 1:
        return []
    
    successors = []
    operators = ['+', '-', '*', '/']
    
    for a, b in permutations(state, 2):
        remaining = [x for x in state if x != a and x != b]  # exclude a and b

        for op in operators:
            result = apply_op(a, b, op)
            if result is not None:
                successors.append(remaining + [result])
    
    return successors
\end{minted}

\vspace{0.2cm}

{\em User}:
Provide only the python code that tests whether a state is a goal state.

\vspace{0.2cm}

{\em ChatGPT}:

\begin{minted}{python}
def is_goal_state(state):
    return len(state) == 1 and abs(state[0] - 24) < 1e-6
\end{minted}

{\em User}:
In the generate\_successor\_states function, two of the operations are not symmetric, division and subtraction. Please fix the code

\vspace{0.2cm}

{\em ChatGPT}:
Here's the updated Python code for generating successor states that considers the non-symmetry of subtraction and division operations:

\begin{minted}{python}
from itertools import permutations, product

def apply_op(a, b, op):
    if op == '+':
        return a + b
    elif op == '-':
        return a - b
    elif op == '*':
        return a * b
    elif op == '/':
        if b != 0:
            return a / b
        else:
            return None

def generate_successor_states(state):
    if len(state) == 1:
        return []
    
    successors = []
    operators = ['+', '-', '*', '/']
    
    # Iterate over all unique pairs of indices to handle non-commutative operations
    n = len(state)
    for i in range(n):
        for j in range(n):
            if i != j:
                a, b = state[i], state[j]
                remaining = [state[k] for k in range(n) if k != i and k != j]

                for op in operators:
                    result = apply_op(a, b, op)
                    if result is not None:
                        new_state = remaining + [result]
                        successors.append(new_state)

                    # Add the reverse operation if it is '-' or '/'
                    if op in {'-', '/'}:
                        result = apply_op(b, a, op)
                        if result is not None:
                            new_state = remaining + [result]
                            successors.append(new_state)

    return successors
\end{minted}

This code correctly handles the non-commutativity of subtraction and division by explicitly computing the result for both $a\circ b$ and $b \circ a$ whenever the operator is '-' or '/'.
\subsubsection{Experiment Run 3.}
{\em User}:
The 24 Game is a mathematical card game in which the objective is to find a way to manipulate four integers so that the end result is 24. The game is played with a list of four numbers, and the player must use all four numbers exactly once, using any combination of addition, subtraction, multiplication, or division, to arrive at the number 24. If a state is defined by a list of numbers (4 or less), how are successor states defined? Provide only the python code that returns a list of successor states for a state. 

\vspace{0.2cm}

{\em ChatGPT}:
Below is the Python code that defines how to generate the successor states for a given state in the 24 Game. A state in the 24 Game is a list of numbers. The successor states are generated by taking two numbers from the state, applying an operation (addition, subtraction, multiplication, or division if the divisor isn't zero), and then forming a new list with the result and the remaining unused numbers.

\begin{minted}{python}
import itertools

def generate_successor_states(state):
    successors = []
    n = len(state)
    for (i, j) in itertools.combinations(range(n), 2):
        a, b = state[i], state[j]
        remaining = [state[k] for k in range(n) if k != i and k != j]
        
        # Generate all possible new states
        successors.append(remaining + [a + b])
        successors.append(remaining + [a * b])
        if a > b:
            successors.append(remaining + [a - b])
        else:
            successors.append(remaining + [b - a])
        if b != 0:
            successors.append(remaining + [a / b])
        if a != 0:
            successors.append(remaining + [b / a])
    
    return successors
\end{minted}

\vspace{0.1cm}
This function takes a state (a list of numbers), iterates over all pairs of numbers, applies the operations to these pairs, and then forms new lists with the result of the operation and the other numbers that were not used in the operation. This ensures that each successor state is a valid step towards finding a solution to reach 24, using all four numbers exactly once along with any of the allowed operations.

\vspace{0.2cm}

{\em User}:
Provide only the python code that tests whether a state is a goal state.

\vspace{0.2cm}

{\em ChatGPT}:
Here's the Python code that tests whether a given state is a goal state in the 24 Game. A goal state is achieved if the list contains exactly one number, and that number is 24.

\begin{minted}{python}
def is_goal_state(state):
    return len(state) == 1 and state[0] == 24
\end{minted}

This function checks if the length of the state list is 1 and whether the sole element in the list is the number 24.
\subsubsection{Experiment Run 4.}
{\em User}:
The 24 Game is a mathematical card game in which the objective is to find a way to manipulate four integers so that the end result is 24. The game is played with a list of four numbers, and the player must use all four numbers exactly once, using any combination of addition, subtraction, multiplication, or division, to arrive at the number 24. If a state is defined by a list of numbers (4 or less), how are successor states defined? Provide only the python code that returns a list of successor states for a state. 

\vspace{0.2cm}

{\em ChatGPT}:
In the 24 Game, the concept of a "state" can be represented by a list of numbers, with the game beginning with four numbers. The "successor states" are generated by applying a binary operation (addition, subtraction, multiplication, division) to any pair of numbers in the current state, then replacing these two numbers with the result of the operation, which gives a new state with one fewer number. Here's Python code that defines a function to generate all possible successor states from a given state:

\begin{minted}{python}
from itertools import combinations, product

def generate_successor_states(numbers):
    def apply_operation(x, y):
        results = {x + y, x - y, y - x, x * y}
        if x != 0:
            results.add(y / x)
        if y != 0:
            results.add(x / y)
        return results

    successors = []
    num_len = len(numbers)
    for (i, j) in combinations(range(num_len), 2):
        for result in apply_operation(numbers[i], numbers[j]):
            new_state = [numbers[k] for k in range(num_len) if k != i and k != j]
            new_state.append(result)
            successors.append(new_state)
    
    return successors
\end{minted}

\vspace{0.1cm}
This code leverages combinations from the itertools module to select each pair of numbers from the current state, and then applies every possible operation to these numbers. The apply\_operation function computes possible outcomes for the operations between two numbers, including division only when the divisor is not zero. Each result is used to create a new state, which is added to the list of successor states.

\vspace{0.2cm}

{\em User}:
Provide only the python code that tests whether a state is a goal state.

\vspace{0.2cm}

{\em ChatGPT}:
Here's Python code that defines a function to check whether a given state is a goal state in the 24 Game. The goal state is when the list contains only the number 24.

\begin{minted}{python}
def is_goal_state(state):
    return len(state) == 1 and state[0] == 24
\end{minted}

This function returns True if the list state contains exactly one element and that element is 24, which would indicate that the goal of achieving 24 has been successfully reached.
\subsubsection{Experiment Run 5.}
{\em User}:
The 24 Game is a mathematical card game in which the objective is to find a way to manipulate four integers so that the end result is 24. The game is played with a list of four numbers, and the player must use all four numbers exactly once, using any combination of addition, subtraction, multiplication, or division, to arrive at the number 24. If a state is defined by a list of numbers (4 or less), how are successor states defined? Provide only the python code that returns a list of successor states for a state. 

\vspace{0.2cm}

{\em ChatGPT}:
To generate successor states for a given state in the 24 Game, we need to consider pairs of numbers from the current state and all possible operations between them. After selecting a pair of numbers and an operation, we replace the pair in the current state with the result of the operation to create a new state. This process is repeated for every possible pair and operation.

Here's a Python function that returns all successor states for a given state:

\begin{minted}{python}
from itertools import combinations

def successor_states(state):
    def apply_operation(a, b):
        # Generate all possible results of operations on a and b
        results = [a + b, a - b, b - a, a * b]
        if b != 0:
            results.append(a / b)
        if a != 0:
            results.append(b / a)
        return results

    successors = []
    n = len(state)
    for i, j in combinations(range(n), 2):
        # Get all possible results of operations between state[i] and state[j]
        for result in apply_operation(state[i], state[j]):
            # Form a new state replacing state[i] and state[j] with the result
            new_state = [state[k] for k in range(n) if k != i and k != j]
            new_state.append(result)
            successors.append(new_state)
    
    return successors
\end{minted}

\vspace{0.1cm}
This function works as follows:

\begin{itemize}
    \item It iterates over all combinations of two indices from the state.
    \item For each pair, it applies all possible arithmetic operations.
    \item It forms new states by replacing the chosen pair with the result of the operation and includes all other numbers from the original state.
    \item It collects and returns all these new states as potential successors.
\end{itemize}

\vspace{0.2cm}

{\em User}:
Provide only the python code that tests whether a state is a goal state.

\vspace{0.2cm}

{\em ChatGPT}:
Here's a Python function to check if a given state is a goal state in the 24 Game. The goal is achieved if the state contains exactly one number which is 24:

\begin{minted}{python}
def is_goal_state(state):
    return len(state) == 1 and state[0] == 24
\end{minted}

This function checks whether the list state has only one element and that element is 24, indicating that the operations on the initial numbers have successfully resulted in 24.

\subsection{Mini crossword}
The prompt for the successor function is as follows.

{\em 
The task is a 5x5 mini crosswords. Given an input of possible answers to horizontal clues and vertical clues, come up with a successor state function written in python, where a state is a 2D array representing the current puzzle state (initially all empty).
}

The prompt for the goal test is as follows.

{\em Can you come up with a goal test function ensuring that all cells are not None and that all words horizontally and vertically match at least one of the answers?
}

\subsubsection{Experiment Run 1.}
{\em User}:
The task is a 5x5 mini crosswords. Given an input of possible answers to horizontal clues and vertical clues, come up with a successor state function written in python, where a state is a 2D array representing the current puzzle state (initially all None). The possible clues for each row and each column are given separately. Here is an example:

\begin{scriptsize}

\begin{minted}{python}
horizontal_answers = [
        ["tasks", "goals", "plans", "agend", "chores", "works", "deeds", "items", "lists", "brief"],
        ["motor", "power", "drive", "diesel", "steam", "pumps", "crank", "gears", "turbn", "motor"],
        ["grand", "artsy", "showy", "ornate", "fancy", "vain", "proud", "vogue", "swank", "luxus"], 
        ["venue", "salle", "forum", "atria", "lobby", "parls", "court", "malls", "mall", "lobby"], 
        ["jeer", "scoff", "sleer", "deris", "sneer", "scorn", "derid", "gibes", "gibed", "flout"] 
]

vertical_answers = [
        ["amass", "stack", "hoard", "pile", "store", "heaps", "massy", "gathe", "lumps", "mound"],
        ["nilga", "goral", "eland", "lepus", "gazal", "kudu", "oryx", "gnu", "imps", "carb"],
        ["scheme", "design", "ettle", "nettle", "sting", "wiles", "plans", "ideas", "plots", "cocks"], 
        ["spout", "nosle", "snout", "mouth", "nostr", "ports", "inlet", "vents", "outlt", "beaks"], 
        ["drier", "arid", "sere", "parch", "dryer", "wring", "drear", "sear", "pall", "lack"]
]
\end{minted}
\end{scriptsize}

\vspace{0.2cm}

{\em ChatGPT}:
To create a successor state function for a 5x5 mini crossword puzzle in Python, we'll consider the following:

\begin{enumerate}
    \item State Representation: A state will be represented by a 5x5 grid, where each cell can either contain a letter or be empty (represented by None or an empty string).
    \item Input: The inputs to the function will be a list of possible answers for horizontal and vertical clues. Each answer is exactly 5 characters long since it's a 5x5 grid.
    \item Successor State Function: This function generates all possible successor states from the current state. For each empty row or column, the function will try to place every suitable word from the clues into the grid and generate a new state for each valid placement.
\end{enumerate}

Here is a Python implementation of the successor state function for the 5x5 mini crossword puzzle:

\begin{minted}{python}
def is_valid_placement(grid, word, start_row, start_col, is_horizontal):
    """
    Checks if a word can be placed at the specified starting row and column.
    """
    for i in range(5):
        if is_horizontal:
            # Check horizontal placement
            if grid[start_row][start_col + i] not in (None, word[i]):
                return False
        else:
            # Check vertical placement
            if grid[start_row + i][start_col] not in (None, word[i]):
                return False
    return True

def place_word(grid, word, start_row, start_col, is_horizontal):
    """
    Places a word on the grid at the specified position.
    """
    new_grid = [row[:] for row in grid]  # Make a copy of the grid
    for i in range(5):
        if is_horizontal:
            new_grid[start_row][start_col + i] = word[i]
        else:
            new_grid[start_row + i][start_col] = word[i]
    return new_grid

def successor_states(grid, horizontal_answers, vertical_answers):
    """
    Generates all successor states for the current state of the grid.
    """
    successors = []

    # Attempt to place words from horizontal_answers in each row
    for row in range(5):
        if all(cell is None for cell in grid[row]):  # If row is completely empty
            for word in horizontal_answers[row]:
                if is_valid_placement(grid, word, row, 0, True):
                    new_grid = place_word(grid, word, row, 0, True)
                    successors.append(new_grid)

    # Attempt to place words from vertical_answers in each column
    for col in range(5):
        if all(grid[row][col] is None for row in range(5)):  # If column is completely empty
            for word in vertical_answers[col]:
                if is_valid_placement(grid, word, 0, col, False):
                    new_grid = place_word(grid, word, 0, col, False)
                    successors.append(new_grid)

    return successors

# Example usage
initial_grid = [[None]*5 for _ in range(5)]
\end{minted}

\begin{scriptsize}
\begin{minted}{python}
horizontal_answers = [
    ["tasks", "goals", "plans", "agend", "chores", "works", "deeds", "items", "lists", "brief"],
    ["motor", "power", "drive", "diesel", "steam", "pumps", "crank", "gears", "turbn", "motor"],
    ["grand", "artsy", "showy", "ornate", "fancy", "vain", "proud", "vogue", "swank", "luxus"],
    ["venue", "salle", "forum", "atria", "lobby", "parls", "court", "malls", "mall", "lobby"],
    ["jeer", "scoff", "sleer", "deris", "sneer", "scorn", "derid", "gibes", "gibed", "flout"]
]
vertical_answers = [
    ["amass", "stack", "hoard", "pile", "store", "heaps", "massy", "gathe", "lumps", "mound"],
    ["nilga", "goral", "eland", "lepus", "gazal", "kudu", "oryx", "gnu", "imps", "carb"],
    ["scheme", "design", "ettle", "nettle", "sting", "wiles", "plans", "ideas", "plots", "cocks"],
    ["spout", "nosle", "snout", "mouth", "nostr", "ports", "inlet", "vents", "outlt", "beaks"],
    ["drier", "arid", "sere", "parch", "dryer", "wring", "drear", "sear", "pall", "lack"]
]
\end{minted}
\end{scriptsize}

\begin{minted}{python}
# Generate successors from the initial state
successors = successor_states(initial_grid, horizontal_answers, vertical_answers)
print("Number of successor states:", len(successors))
for state in successors:
    for row in state:
        print(row)
    print()

\end{minted}

Explanation
\begin{itemize}
    \item is\_valid\_placement: Checks if the word can be placed on the grid without conflicting with existing characters.
    \item place\_word: Places the word on the grid and returns the new grid state.
    \item successor\_states: Iterates over each row and column to find empty spaces and attempts to place each clue word, generating new successor states.
\end{itemize}
This function will give you all the potential successor states by attempting to fill each empty row and column with every possible word from the clues. Note that this example assumes the initial state has no conflicting pre-filled characters and does not handle partial word placements or overlaps, which could be extensions to this basic successor function.

\vspace{0.2cm}

{\em User}:
Can you come up with a goal test function Come up with a goal test function ensuring that all cells are not None and that all words horizontally and vertically match at least one of the answers?

\vspace{0.2cm}

{\em ChatGPT}:
Certainly! To create a goal test function for the 5x5 mini crossword puzzle, we need to ensure that:

\begin{enumerate}
    \item No Empty Cells: There should be no cell in the grid that contains None.
    \item Valid Words Horizontally: Each row must contain a word that is present in the list of possible horizontal answers for that specific row.
    \item Valid Words Vertically: Each column must contain a word that is present in the list of possible vertical answers for that specific column.
\end{enumerate}
Here is how you could implement such a function in Python:

\begin{minted}{python}
def goal_test(grid, horizontal_answers, vertical_answers):
    """
    Check if the current grid state is a goal state.
    """
    # Check for any None values in the grid
    if any(None in row for row in grid):
        return False

    # Check horizontal words against possible answers
    for row in range(5):
        row_word = ''.join(grid[row])
        if row_word not in horizontal_answers[row]:
            return False

    # Check vertical words against possible answers
    for col in range(5):
        col_word = ''.join(grid[row][col] for row in range(5))
        if col_word not in vertical_answers[col]:
            return False

    return True

# Example of usage
\end{minted}
\begin{scriptsize}

\begin{minted}{python}
horizontal_answers = [
    ["tasks", "goals", "plans", "agend", "chores", "works", "deeds", "items", "lists", "brief"],
    ["motor", "power", "drive", "diesel", "steam", "pumps", "crank", "gears", "turbn", "motor"],
    ["grand", "artsy", "showy", "ornate", "fancy", "vain", "proud", "vogue", "swank", "luxus"],
    ["venue", "salle", "forum", "atria", "lobby", "parls", "court", "malls", "mall", "lobby"],
    ["jeer", "scoff", "sleer", "deris", "sneer", "scorn", "derid", "gibes", "gibed", "flout"]
]
vertical_answers = [
    ["amass", "stack", "hoard", "pile", "store", "heaps", "massy", "gathe", "lumps", "mound"],
    ["nilga", "goral", "eland", "lepus", "gazal", "kudu", "oryx", "gnu", "imps", "carb"],
    ["scheme", "design", "ettle", "nettle", "sting", "wiles", "plans", "ideas", "plots", "cocks"],
    ["spout", "nosle", "snout", "mouth", "nostr", "ports", "inlet", "vents", "outlt", "beaks"],
    ["drier", "arid", "sere", "parch", "dryer", "wring", "drear", "sear", "pall", "lack"]
]

\end{minted}
\end{scriptsize}

\begin{minted}{python}
# Assume a sample grid which is a valid goal state
sample_grid = [
    ["t", "a", "s", "k", "s"],
    ["m", "o", "t", "o", "r"],
    ["s", "w", "a", "n", "k"],
    ["m", "a", "l", "l", "s"],
    ["d", "e", "r", "i", "s"]
]

# Test the goal test function
is_goal = goal_test(sample_grid, horizontal_answers, vertical_answers)
print("Is this a goal state?", is_goal)
\end{minted}

Function Explanation:
\begin{itemize}
    \item The goal\_test function iterates over all rows and columns. For rows, it concatenates the elements to form a word and checks against the list of possible answers for that row. It repeats a similar process for columns. If any word doesn't match the possible answers or if there's any None value in the grid, the function returns False.
    \item If all checks pass, it returns True, indicating that the grid is in a goal state.
\end{itemize}

\vspace{0.2cm}

{\em User}:

For the successor state generation I am getting the following error:
\begin{minted}{pycon}
Traceback (most recent call last):
  File "./gpt4_successors1.py", line 70, in <module>
    successors = successor_states(initial_grid, horizontal_answers, vertical_answers)
  File "./gpt4_successors1.py", line 38, in successor_states
    if is_valid_placement(grid, word, row, 0, True):
  File "./gpt4_successors1.py", line 8, in is_valid_placement
    if grid[start_row][start_col + i] not in (None, word[i]):
IndexError: string index out of range
\end{minted}

\vspace{0.2cm}

{\em ChatGPT}:
The error message "IndexError: string index out of range" typically indicates an attempt to access an index that does not exist in a string. This issue often arises when working with grid-related operations and is usually due to one of two reasons in your crossword context:

\begin{enumerate}
    \item Mismatch in Word Length: Since your grid cells are designed for 5x5 crosswords, every word used as a clue should be exactly 5 characters long. If any word is shorter or longer than this, you'll encounter this error when you try to place it in the grid.
    \item Out-of-Bounds Access: This happens if the code attempts to place a word in a position where it doesn't fit entirely within the grid's boundaries, although based on your function setup (starting placements at row 0 and column 0 for horizontal and vertical placements respectively) and the checks you seem to have intended, this shouldn't normally occur unless the word length is incorrect.
\end{enumerate}

How to Fix:
\begin{enumerate}
    \item Verify Word Lengths: Ensure that all words provided in the horizontal\_answers and vertical\_answers lists are exactly 5 characters long. You can add a validation step or filter out any incorrect entries.
    \item Debugging Additions: Include checks in your is\_valid\_placement function to ensure you don't run into out-of-bounds errors. This is more of a safeguard than a solution to the current issue but is good practice.
\end{enumerate}

Here’s how you might modify your code to include these fixes:

\begin{minted}{python}
def is_valid_placement(grid, word, start_row, start_col, is_horizontal):
    """
    Checks if a word can be placed at the specified starting row and column.
    """
    if len(word) != 5:
        # Ensure the word is exactly 5 characters as expected for a 5x5 grid
        return False  
        
    try:
        for i in range(5):
            if is_horizontal:
                # Check horizontal placement
                if grid[start_row][start_col + i] not in (None, word[i]):
                    return False
            else:
                # Check vertical placement
                if grid[start_row + i][start_col] not in (None, word[i]):
                    return False
        return True
    except IndexError:
        return False  # Safeguard against out-of-bounds access

# Additionally, add a check when preparing data
horizontal_answers = [
    [word for word in lst if len(word) == 5] for lst in horizontal_answers
]
vertical_answers = [
    [word for word in lst if len(word) == 5] for lst in vertical_answers
]
\end{minted}
Verify the Data:
Before running your grid operations, it's beneficial to confirm that your input data is as expected:

\begin{minted}{python}
# Example of verifying the word lengths
all(len(word) == 5 for row in horizontal_answers for word in row) and \
all(len(word) == 5 for col in vertical_answers for word in col)
\end{minted}
This code snippet ensures that every word in both the horizontal and vertical clues lists is exactly 5 characters long. If this assertion fails, you will need to revise your input data to meet this requirement. This helps prevent runtime errors related to string length mismatches in a fixed-size grid context like this.

\subsubsection{Experiment Run 2.}
{\em User}:
The task is a 5x5 mini crosswords. Given an input of possible answers to horizontal clues and vertical clues, come up with a successor state function written in python, where a state is a 2D array representing the current puzzle state (initially all None). The possible clues for each row and each column are given separately. Here is an example:

\begin{scriptsize}
\begin{minted}{python}
horizontal_answers = [
        ["tasks", "goals", "plans", "agend", "chores", "works", "deeds", "items", "lists", "brief"],
        ["motor", "power", "drive", "diesel", "steam", "pumps", "crank", "gears", "turbn", "motor"],
        ["grand", "artsy", "showy", "ornate", "fancy", "vain", "proud", "vogue", "swank", "luxus"], 
        ["venue", "salle", "forum", "atria", "lobby", "parls", "court", "malls", "mall", "lobby"], 
        ["jeer", "scoff", "sleer", "deris", "sneer", "scorn", "derid", "gibes", "gibed", "flout"] 
]

vertical_answers = [
        ["amass", "stack", "hoard", "pile", "store", "heaps", "massy", "gathe", "lumps", "mound"],
        ["nilga", "goral", "eland", "lepus", "gazal", "kudu", "oryx", "gnu", "imps", "carb"],
        ["scheme", "design", "ettle", "nettle", "sting", "wiles", "plans", "ideas", "plots", "cocks"], 
        ["spout", "nosle", "snout", "mouth", "nostr", "ports", "inlet", "vents", "outlt", "beaks"], 
        ["drier", "arid", "sere", "parch", "dryer", "wring", "drear", "sear", "pall", "lack"]
]
\end{minted}
\end{scriptsize}

Note that some of the possible answers are not exactly 5 character long. 
\vspace{0.2cm}

{\em ChatGPT}:
To approach the creation of a successor state function for this 5x5 mini crossword problem in Python, let's consider a few key aspects:

\begin{enumerate}
    \item State Representation: We'll represent the state of the puzzle as a 2D list of characters. Each cell in the list can either be a specific character or empty (denoted by a space or some placeholder).
    \item Initial State: The initial state will have all cells empty.
    \item Successor State Function: This function will generate all possible states that can be reached from the current state by making a single move. A "move" in this context would involve placing a word in a row or column that fits with the already placed characters.
    \item Constraints:
    \begin{itemize}
        \item Words must match the allowed words in horizontal\_answers for rows and vertical\_answers for columns.
        \item Any placed word must not conflict with characters already placed in intersecting rows or columns.
    \end{itemize}
\end{enumerate}

Given the constraints and specifications, we'll proceed as follows:
\begin{itemize}
    \item Successor State Function: For a given state, iterate over each row and column, trying to place each possible word from the respective lists. Generate a new state for each valid placement.
\end{itemize}

Here's a simplified Python implementation of this successor state function:

\begin{minted}{python}
def is_valid_word_placement(grid, word, row=None, col=None):
    """
    Check if a word can be placed at the specified row or column.
    """
    n = len(grid)
    if row is not None:
        for j in range(n):
            if grid[row][j] is not None and grid[row][j] != word[j]:
                return False
    elif col is not None:
        for i in range(n):
            if grid[i][col] is not None and grid[i][col] != word[i]:
                return False
    return True

def place_word(grid, word, row=None, col=None):
    """
    Place a word at the specified row or column.
    """
    new_grid = [row[:] for row in grid]  # Copy the grid
    if row is not None:
        new_grid[row] = list(word)
    elif col is not None:
        for i in range(len(word)):
            new_grid[i][col] = word[i]
    return new_grid

def generate_successors(state, horizontal_answers, vertical_answers):
    """
    Generate all possible successor states from the current state.
    """
    n = len(state)
    successors = []

    # Try placing each word in each row
    for i in range(n):
        for word in horizontal_answers[i]:
            if is_valid_word_placement(state, word, row=i):
                new_state = place_word(state, word, row=i)
                successors.append(new_state)

    # Try placing each word in each column
    for j in range(n):
        for word in vertical_answers[j]:
            if is_valid_word_placement(state, word, col=j):
                new_state = place_word(state, word, col=j)
                successors.append(new_state)

    return successors

# Example usage
initial_state = [[None]*5 for _ in range(5)]
successors = generate_successors(initial_state, horizontal_answers, vertical_answers)

# Print first few successor states for demonstration
for s in successors[:3]:
    for row in s:
        print(''.join(row))
    print("---")
\end{minted}
This code assumes the answers are all 5 characters long to match the 5x5 grid exactly. For words that are not 5 characters, additional preprocessing would be necessary to either filter them out or adjust the grid size accordingly. Also, this code only illustrates the initial placement of words without handling intersections; further logic is needed to ensure words do not contradict each other at intersecting cells.

\vspace{0.2cm}

{\em User}:
Can you write a goal test as well?
\vspace{0.2cm}

{\em ChatGPT}:
Certainly! A goal test function for the 5x5 mini crossword puzzle would determine if a complete and valid crossword has been formed. Here are the criteria for a complete and valid crossword:
\begin{enumerate}
    \item Full Grid: All cells in the grid must be filled with valid letters (no empty spaces allowed).
    \item Valid Words: All words formed in rows and columns must match one of the possible answers given for that specific row or column.
\end{enumerate}
Let's write a Python function that checks if a given state meets these criteria. This goal test function will:
\begin{itemize}
    \item Verify that all grid cells are filled.
    \item Check that every word in each row is in the respective list from horizontal\_answers.
    \item Check that every word in each column is in the respective list from vertical\_answers.
\end{itemize}
Here is the implementation:

\begin{minted}{python}
def is_complete(state):
    """
    Check if all cells in the state are filled with non-space characters.
    """
    return all(state[i][j] is not None for i in range(len(state)) for j in range(len(state[i])))

def is_valid_solution(state, horizontal_answers, vertical_answers):
    """
    Check if the current state is a complete and valid solution.
    """
    n = len(state)
    
    # Check if all rows contain valid words
    for i in range(n):
        row_word = ''.join(state[i])
        if row_word not in horizontal_answers[i]:
            return False

    # Check if all columns contain valid words
    for j in range(n):
        col_word = ''.join(state[i][j] for i in range(n))
        if col_word not in vertical_answers[j]:
            return False

    return True

def goal_test(state, horizontal_answers, vertical_answers):
    """
    Determine if the given state is a goal state.
    """
    if not is_complete(state):
        return False
    return is_valid_solution(state, horizontal_answers, vertical_answers)

# Example usage
# Assume `final_state` is a hypothetical fully filled grid you want to test
final_state = [
    ['t', 'a', 's', 'k', 's'],
    ['m', 'o', 't', 'o', 'r'],
    ['g', 'r', 'a', 'n', 'd'],
    ['v', 'e', 'n', 'u', 'e'],
    ['j', 'e', 'e', 'r', 's']
]

# Check if the final_state is a goal state
is_goal = goal_test(final_state, horizontal_answers, vertical_answers)
print("Is the final state a goal?", is_goal)
\end{minted}

This goal\_test function relies on final\_state being fully filled and accurately matching the corresponding answer lists. This function can be used in a puzzle-solving algorithm to check if the current state of the board is the solution.

\vspace{0.2cm}

{\em User}:
I am getting an error 
\begin{minted}{pycon}
"new_grid[i][col] = word[i]
IndexError: list index out of range"
\end{minted}
\vspace{0.2cm}

{\em ChatGPT}:
The error you're experiencing suggests that there's a mismatch between the indices being accessed and the size of the grid or word. This usually happens if the grid is initialized incorrectly, or if the word doesn't fit exactly into the intended row or column. Since the code assumes each word perfectly fits a 5x5 grid (5 letters per word), any deviation from this can cause an index out of range error.

Let's address this by ensuring that:
\begin{enumerate}
    \item The grid and the words are correctly aligned in size.
    \item The placement functions are handling indexing safely.
\end{enumerate}

I'll revise the place\_word function to ensure it correctly handles the dimensions of both rows and columns:

\begin{minted}{python}
def place_word(grid, word, row=None, col=None):
    """
    Place a word at the specified row or column.
    """
    new_grid = [list(r) if r is not None else [None]*5 for r in grid]  # Safely copy the grid
    n = len(grid)
    if row is not None:
        if len(word) != n:
            # Return original grid if word length does not match, preventing index error
            return grid  
        new_grid[row] = list(word)
    elif col is not None:
        if len(word) != n:
            # Return original grid if word length does not match, preventing index error
            return grid  
        for i in range(n):
            new_grid[i][col] = word[i]
    return new_grid
\end{minted}
This version of place\_word adds checks to ensure that words are only placed if they match the expected length, preventing indexing errors.

Additionally, ensure that the initial grid is correctly populated with None values in each cell:

\begin{minted}{python}
initial_state = [[None]*5 for _ in range(5)]
\end{minted}
If you are still encountering issues, please double-check that:

\begin{itemize}
    \item All words used in horizontal\_answers and vertical\_answers are exactly 5 characters long.
    \item The grid dimensions are strictly 5x5.
\end{itemize}
If you need to handle words that aren't 5 characters long, you'd need additional logic to either skip those words or adjust the grid accordingly. If you're working with dynamic sizes or irregular word lengths, the implementation would require more complex adjustments to manage these variations.

\vspace{0.2cm}

{\em User}:
I am now getting this error: 
\begin{minted}{pycon}
File "./gpt4_successors2.py", line 13, in is_valid_word_placement
    if grid[i][col] is not None and grid[i][col] != word[i]:
IndexError: string index out of range
\end{minted}
\vspace{0.2cm}

{\em ChatGPT}:
The error "IndexError: string index out of range" in the is\_valid\_word\_placement function suggests that there's an attempt to access an index in the word that is beyond its length. This can occur if:

\begin{enumerate}
    \item The length of the word being checked does not match the expected column height (which should be 5 for your 5x5 grid).
    \item The column index i exceeds the length of the word.
\end{enumerate}
To resolve this issue, we should verify that:

\begin{itemize}
    \item All words used for placement in rows and columns are exactly 5 characters long (as your grid is a 5x5 grid).
    \item The checks within is\_valid\_word\_placement are safely accessing characters in words.
\end{itemize}
Here's an updated version of the is\_valid\_word\_placement function with added safety checks:

\begin{minted}{python}
def is_valid_word_placement(grid, word, row=None, col=None):
    """
    Check if a word can be placed at the specified row or column.
    Ensure the word is the correct length and fits the existing grid values.
    """
    n = len(grid)
    if len(word) != n:
        return False  # The word does not fit the expected dimension

    if row is not None:
        for j in range(n):
            if grid[row][j] is not None and grid[row][j] != word[j]:
                return False
    elif col is not None:
        for i in range(n):
            if grid[i][col] is not None and grid[i][col] != word[i]:
                return False
    return True
\end{minted}

This modified function includes a check to ensure the word length matches the grid dimension (5 for your 5x5 grid) before proceeding with any further logic. This should help prevent the "IndexError" you encountered by immediately returning False if the word length is incorrect.

Please ensure that:

\begin{itemize}
    \item All words in horizontal\_answers and vertical\_answers are exactly 5 characters before using them in your code.
    \item The place\_word function is also safely handling word placements, particularly in the columns.
\end{itemize}

These changes should help eliminate the index errors by ensuring all accessed indices are within the valid range for both the grid and the words. If you have any additional needs or run into further issues, feel free to ask for more assistance.

\vspace{0.2cm}


\subsubsection{Experiment Run 3.}

{\em User}:
The task is a 5x5 mini crosswords. Given an input of possible answers to horizontal clues and vertical clues, come up with a successor state function written in python, where a state is a 2D array representing the current puzzle state (initially all None). The possible clues for each row and each column are given separately. Here is an example:

\begin{scriptsize}
\begin{minted}{python}
horizontal_answers = [
        ["tasks", "goals", "plans", "agend", "chores", "works", "deeds", "items", "lists", "brief"],
        ["motor", "power", "drive", "diesel", "steam", "pumps", "crank", "gears", "turbn", "motor"],
        ["grand", "artsy", "showy", "ornate", "fancy", "vain", "proud", "vogue", "swank", "luxus"], 
        ["venue", "salle", "forum", "atria", "lobby", "parls", "court", "malls", "mall", "lobby"], 
        ["jeer", "scoff", "sleer", "deris", "sneer", "scorn", "derid", "gibes", "gibed", "flout"] 
]

vertical_answers = [
        ["amass", "stack", "hoard", "pile", "store", "heaps", "massy", "gathe", "lumps", "mound"],
        ["nilga", "goral", "eland", "lepus", "gazal", "kudu", "oryx", "gnu", "imps", "carb"],
        ["scheme", "design", "ettle", "nettle", "sting", "wiles", "plans", "ideas", "plots", "cocks"], 
        ["spout", "nosle", "snout", "mouth", "nostr", "ports", "inlet", "vents", "outlt", "beaks"], 
        ["drier", "arid", "sere", "parch", "dryer", "wring", "drear", "sear", "pall", "lack"]
]
\end{minted}
\end{scriptsize}

Note that some of the possible answers are not exactly 5 character long. 
\vspace{0.2cm}

\vspace{0.2cm}

{\em ChatGPT}:
To address the mini crossword puzzle problem as described, we need to define a successor function for generating possible states from a given state. Each state in this problem is represented by a 5x5 grid, where each cell can contain a letter or be empty.

Here's a high-level plan for the successor function:

\begin{enumerate}
    \item Initialize State: Start with an empty 5x5 grid.
\item  Identify Next Cell to Fill: Decide on a strategy to fill the grid. A simple approach is row-wise from top to bottom, left to right.
\item  Generate Successors: For the next empty cell, consider all possible words that could fit in the row and column intersecting at that cell, ensuring that the words do not conflict with already filled cells.
\item  Check Validity: Each attempted word must not only fit the current cell but also not contradict any existing letters in intersecting rows and columns.
\item Create New State: For each valid placement, create a new grid state.
\end{enumerate}

Below is a Python function to implement this idea. It uses recursion to try placing words in the grid row by row and backtracks when no valid placement is possible. It assumes that the lists of words contain only valid 5-character words fitting exactly in the grid. We'll have to pre-process the input to ensure this:

\begin{minted}{python}
def is_valid(grid, row, col, word, vertical=False):
    """ Check if placing the `word` at `row`, `col` is valid in the grid. """
    if vertical:
        for i in range(5):
            # Check if the cell is not None and does not match the word character
            if grid[row + i][col] is not None and grid[row + i][col] != word[i]:
                return False
    else:
        for i in range(5):
            # Check if the cell is not None and does not match the word character
            if grid[row][col + i] is not None and grid[row][col + i] != word[i]:
                return False
    return True

def place_word(grid, row, col, word, vertical=False):
    """ Place the `word` at `row`, `col` in the grid. """
    new_grid = [list(r) for r in grid]  # Create a deep copy of the grid
    if vertical:
        for i in range(5):
            new_grid[row + i][col] = word[i]
    else:
        for i in range(5):
            new_grid[row][col + i] = word[i]
    return new_grid

def successor(grid, horizontal_answers, vertical_answers):
    """ Generate successors by trying to place words into the grid. """
    empty_row, empty_col = -1, -1
    for r in range(5):
        for c in range(5):
            if grid[r][c] is None:
                empty_row, empty_col = r, c
                break
        if empty_row != -1:
            break

    if empty_row == -1:  # If no empty cell found, the grid is complete
        return [grid]

    successors = []
    # Try to place horizontal words
    for word in horizontal_answers[empty_row]:
        if is_valid(grid, empty_row, empty_col, word):
            successors.append(place_word(grid, empty_row, empty_col, word))

    # Try to place vertical words
    for word in vertical_answers[empty_col]:
        if is_valid(grid, empty_row, empty_col, word, vertical=True):
            successors.append(place_word(grid, empty_row, empty_col, word, vertical=True))

    return successors

# Initialize the grid with None
initial_grid = [[None]*5 for _ in range(5)]

# Example usage
# Suppose we have already preprocessed horizontal_answers and vertical_answers to have valid 5-letter words only.
successors = successor(initial_grid, horizontal_answers, vertical_answers)
print("Number of successors from initial state:", len(successors))
\end{minted}

This function assumes that all entries in horizontal\_answers and vertical\_answers are valid 5-character words. The function is\_valid checks if a word can be placed at a given position without conflicts, and place\_word places the word on the grid. The successor function generates all possible successor states by trying to place words in the first available empty spot.

\vspace{0.2cm}

{\em User}:
Can you write a goal test as well?

\vspace{0.2cm}

{\em ChatGPT}:
A goal test function for the mini crossword puzzle checks if the crossword puzzle is completely filled with valid words according to both the horizontal and vertical constraints. Here’s how you can implement such a function:

\begin{enumerate}
    \item  Check Completion: Verify that all cells in the grid are filled (i.e., none are None).
    \item Validate Words: Ensure that every row and column contains a word from the respective lists of possible words (horizontal\_answers and vertical\_answers).
\end{enumerate}

Here's the Python code that implements the goal test:

\begin{minted}{python}
def is_complete(grid):
    """ Check if all cells in the grid are filled with a character. """
    for row in grid:
        if any(cell is None for cell in row):
            return False
    return True

def extract_word(grid, row, col, vertical=False):
    """ Extract a word starting from (row, col) horizontally or vertically. """
    word = ""
    if vertical:
        for i in range(5):
            word += grid[row + i][col]
    else:
        for i in range(5):
            word += grid[row][col + i]
    return word

def validate_words(grid, answers, vertical=False):
    """ Validate all rows or columns in the grid against the given answers. """
    for i in range(5):
        if vertical:
            word = extract_word(grid, 0, i, vertical=True)
        else:
            word = extract_word(grid, i, 0)
        if word not in answers[i]:
            return False
    return True

def goal_test(grid, horizontal_answers, vertical_answers):
    """ Determine if the current grid state satisfies the goal conditions. """
    if not is_complete(grid):
        return False
    if not validate_words(grid, horizontal_answers, vertical=False):
        return False
    if not validate_words(grid, vertical_answers, vertical=True):
        return False
    return True

# Example usage of the goal test
# Suppose `final_grid` is a grid state you want to check if it meets the goal conditions.
# final_grid would typically be a completed grid you're checking for correctness.
# goal_result = goal_test(final_grid, horizontal_answers, vertical_answers)
# print("Does the final grid meet the goal conditions?", goal_result)
\end{minted}

Explanation:

\begin{itemize}
\item is\_complete: This function checks if there are any None cells left in the grid. If there are, the grid is not yet complete.
\item  extract\_word: This function extracts a word from a specified starting point either horizontally or vertically.
\item  validate\_words: This function checks each row or column to ensure the extracted word is in the provided list of possible answers for that row or column.
\item goal\_test: This function combines all the checks to determine if the grid meets the goal conditions.
\end{itemize}

This set of functions together forms the complete goal test for the mini crossword problem.

\vspace{0.2cm}

{\em User}:

I am getting this error:
\begin{minted}{python}
File "./gpt4_successors3.py", line 21, in place_word
    new_grid[row + i][col] = word[i]
IndexError: string index out of range
\end{minted}

\vspace{0.2cm}

{\em ChatGPT}:
The "IndexError: string index out of range" error you're encountering suggests that either word[i] or new\_grid[row + i][col] is trying to access an index that doesn't exist. This could be happening because of several reasons in the context of your crossword puzzle:

\begin{enumerate}
    \item Word Length Mismatch: If word is shorter than 5 characters (since you are working with a 5x5 grid), trying to access word[i] where i ranges from 0 to 4 will result in an index error if word is shorter than this range.

    \item  Grid Boundaries: The same error could occur if you try to place a word such that it would extend beyond the boundaries of the grid, but this usually would throw a different kind of error unless the grid's structure itself is misdefined.

\end{enumerate}

Steps to Resolve:
First, ensure that all words used in the successors function are exactly 5 characters long. This needs to be enforced before words are passed into your place\_word and is\_valid functions. You can preprocess your horizontal\_answers and vertical\_answers lists to filter out any words that are not exactly 5 characters long:

\begin{minted}{python}
horizontal_answers = [list(filter(lambda x: len(x) == 5, row)) for row in horizontal_answers]
vertical_answers = [list(filter(lambda x: len(x) == 5, col)) for col in vertical_answers]
\end{minted}

Second, double-check that you are not attempting to place a word vertically or horizontally in a position where it doesn't fit within the grid boundaries. This check should be integrated within your is\_valid function to ensure that you do not start placing a word too close to the edge of the grid:

\begin{minted}{python}
def is_valid(grid, row, col, word, vertical=False):
    """ Check if placing the `word` at `row`, `col` is valid in the grid. """
    if vertical:
        # Ensure the word will not go out of the grid's bounds
        if row + 4 >= 5:
            return False
        for i in range(5):
            if grid[row + i][col] is not None and grid[row + i][col] != word[i]:
                return False
    else:
        # Ensure the word will not go out of the grid's bounds
        if col + 4 >= 5:
            return False
        for i in range(5):
            if grid[row][col + i] is not None and grid[row][col + i] != word[i]:
                return False
    return True
\end{minted}

Implementing these checks should help prevent the IndexError by ensuring that only suitable words are considered and they are always placed within the valid boundaries of the grid. If these solutions do not resolve the issue, I recommend printing out the values of row, col, and word just before the error occurs to better understand the problematic scenario.

\vspace{0.2cm}

{\em User}:

I am still getting this error 

\begin{minted}{python}
"new_grid[row + i][col] = word[i]"
\end{minted}

\vspace{0.2cm}

{\em ChatGPT}:
The error message "IndexError: string index out of range" at new\_grid[row + i][col] = word[i] suggests an issue with accessing an index that doesn't exist, either because the word length is incorrect or because you're trying to access a grid position that's out of range. Since we've already checked the word lengths to ensure they're exactly 5 characters, let's make sure of a few additional things:

Debugging Steps:
\begin{enumerate}

\item Check Grid Initialization: Verify that your grid is indeed a 5x5 grid and initialized correctly. This error might occur if the grid was not initialized as expected.

\item Confirm Word Lengths: Double-check that the word lengths are exactly 5 characters when they are passed into the place\_word function. Even though you've filtered them before, it's good to check right at the point of use.

\item Boundary Conditions in Placement: The error might be occurring when trying to place a word vertically near the bottom of the grid or horizontally near the right edge, even though you have checks to prevent this. Let's add explicit checks right in the place\_word function to avoid going out of bounds.
\end{enumerate}

Here’s how you can add debugging statements and ensure the grid dimensions and word length:

Updated Code with Debugging:

\begin{minted}{python}
def place_word(grid, row, col, word, vertical=False):
    """ Place the `word` at `row`, `col` in the grid. """
    new_grid = [list(r) for r in grid]  # Create a deep copy of the grid
    try:
        if vertical:
            # Additional boundary check to prevent IndexError
            if row + 4 >= len(grid):
                print(f"Trying to place vertically out of bounds: {word} at ({row},{col})")
                return grid
            for i in range(5):
                new_grid[row + i][col] = word[i]
        else:
            # Additional boundary check to prevent IndexError
            if col + 4 >= len(grid[row]):
                print(f"Trying to place horizontally out of bounds: {word} at ({row},{col})")
                return grid
            for i in range(5):
                new_grid[row][col + i] = word[i]
        return new_grid
    except IndexError as e:
        print(f"Error placing word: {word} at ({row},{col}), vertical={vertical}")
        print(f"Grid dimensions: {len(grid)}x{len(grid[0])}")
        print(f"Word length: {len(word)}")
        raise e

# Ensure this function is called with appropriate checking.
# Also, make sure your grid initialization and filtering are correctly applied:
initial_grid = [[None]*5 for _ in range(5)]  # Confirming grid initialization
horizontal_answers = [list(filter(lambda x: len(x) == 5, row)) for row in horizontal_answers]
vertical_answers = [list(filter(lambda x: len(x) == 5, col)) for col in vertical_answers]

# The rest of your implementation
\end{minted}

Explanation:
\begin{itemize}

\item Debug Prints: Added debug print statements that will show exactly what word and position are being attempted if an error occurs or if a boundary check fails.
\item Boundary Checks: Explicit checks are added right in the place\_word function to avoid placing words that would extend beyond the grid's boundaries.
\item Error Handling: If an IndexError is caught, the code will print out detailed debug information before re-raising the exception. This can help identify exactly what went wrong.
\end{itemize}

Using these additions, you should be able to pinpoint why the error occurs. Ensure all words are properly filtered to 5 characters before they are used in the place\_word function, and always check your grid initialization and the indices you're accessing.

\vspace{0.2cm}

{\em User}:
Instead of throwing error, just skip over the words that are not exactly 5 character long and revise the definition.

\vspace{0.2cm}

{\em ChatGPT}:
To avoid errors related to word length mismatches and to make your code more robust, I'll provide you with a revised version of the crossword puzzle functions where any words that are not exactly 5 characters long are skipped during the placement process. This will ensure that only suitable words are considered when generating successor states and checking grid validity.

Here's the updated Python code:

Revised Code to Skip Invalid Length Words and Placement

\begin{minted}{python}
def is_valid(grid, row, col, word, vertical=False):
    """ Check if placing the `word` at `row`, `col` is valid in the grid. """
    if len(word) != 5:  # Skip words that are not exactly 5 characters long
        return False
    if vertical:
        if row + 4 >= 5:
            return False
        for i in range(5):
            if grid[row + i][col] is not None and grid[row + i][col] != word[i]:
                return False
    else:
        if col + 4 >= 5:
            return False
        for i in range(5):
            if grid[row][col + i] is not None and grid[row][col + i] != word[i]:
                return False
    return True

def place_word(grid, row, col, word, vertical=False):
    """ Place the `word` at `row`, `col` in the grid. """
    if len(word) != 5:  # Do not place words that are not exactly 5 characters long
        return grid
    new_grid = [list(r) for r in grid]  # Create a deep copy of the grid
    if vertical:
        for i in range(5):
            new_grid[row + i][col] = word[i]
    else:
        for i in range(5):
            new_grid[row][col + i] = word[i]
    return new_grid

def successor(grid, horizontal_answers, vertical_answers):
    """ Generate successors by trying to place words into the grid. """
    empty_row, empty_col = -1, -1
    for r in range(5):
        for c in range(5):
            if grid[r][c] is None:
                empty_row, empty_col = r, c
                break
        if empty_row != -1:
            break

    if empty_row == -1:  # If no empty cell found, the grid is complete
        return [grid]

    successors = []
    # Try to place horizontal words
    for word in horizontal_answers[empty_row]:
        if is_valid(grid, empty_row, empty_col, word):
            successors.append(place_word(grid, empty_row, empty_col, word))

    # Try to place vertical words
    for word in vertical_answers[empty_col]:
        if is_valid(grid, empty_row, empty_col, word, vertical=True):
            successors.append(place_word(grid, empty_row, empty_col, word, vertical=True))

    return successors

# Initialize the grid with None
initial_grid = [[None]*5 for _ in range(5)]

# Example usage
# Ensure that all words passed to the successor function are exactly 5 characters long
horizontal_answers = [list(filter(lambda x: len(x) == 5, row)) for row in horizontal_answers]
vertical_answers = [list(filter(lambda x: len(x) == 5, col)) for col in vertical_answers]

successors = successor(initial_grid, horizontal_answers, vertical_answers)
print("Number of successors from initial state:", len(successors))
\end{minted}

Key Changes:
\begin{itemize}
    \item  Word Length Check: Before attempting to place a word or validate its placement, the functions now check if the word is exactly 5 characters long. If not, they skip processing that word.
    \item Code Efficiency: Filters are applied to ensure that all words considered are of appropriate length before any operations are performed with them. This helps avoid unnecessary runtime errors and simplifies the logic within the successor functions.
\end{itemize}
This approach ensures your crossword solver doesn't crash due to incorrect word lengths and that it handles words dynamically based on actual content suitability.

\subsubsection{Experiment Run 4.}

{\em User}:
The task is a 5x5 mini crosswords. Given an input of possible answers to horizontal clues and vertical clues, come up with a successor state function written in python, where a state is a 2D array representing the current puzzle state (initially all None). The possible clues for each row and each column are given separately. Here is an example:

\begin{scriptsize}
\begin{minted}{python}
horizontal_answers = [
        ["tasks", "goals", "plans", "agend", "chores", "works", "deeds", "items", "lists", "brief"],
        ["motor", "power", "drive", "diesel", "steam", "pumps", "crank", "gears", "turbn", "motor"],
        ["grand", "artsy", "showy", "ornate", "fancy", "vain", "proud", "vogue", "swank", "luxus"], 
        ["venue", "salle", "forum", "atria", "lobby", "parls", "court", "malls", "mall", "lobby"], 
        ["jeer", "scoff", "sleer", "deris", "sneer", "scorn", "derid", "gibes", "gibed", "flout"] 
]

vertical_answers = [
        ["amass", "stack", "hoard", "pile", "store", "heaps", "massy", "gathe", "lumps", "mound"],
        ["nilga", "goral", "eland", "lepus", "gazal", "kudu", "oryx", "gnu", "imps", "carb"],
        ["scheme", "design", "ettle", "nettle", "sting", "wiles", "plans", "ideas", "plots", "cocks"], 
        ["spout", "nosle", "snout", "mouth", "nostr", "ports", "inlet", "vents", "outlt", "beaks"], 
        ["drier", "arid", "sere", "parch", "dryer", "wring", "drear", "sear", "pall", "lack"]
]
\end{minted}
\end{scriptsize}

Note that some of the possible answers are not exactly 5 character long. 
\vspace{0.2cm}

\vspace{0.2cm}

{\em ChatGPT}:
To solve this problem, we'll need to define a successor function for a 5x5 crossword puzzle using the given possible answers for horizontal and vertical clues. The successor function will produce all possible next states from the current state by placing valid words into the crossword, ensuring that the new word fits both the horizontal and vertical constraints.

Here's a Python function that outlines this process. The function assumes that the state is represented by a 5x5 grid where each cell can contain a letter or be empty (denoted by a space). The function will attempt to place a word from the list of possible answers into the next available row or column that is currently empty, ensuring it does not violate any constraints imposed by already placed letters.

\begin{minted}{python}
def successor(state, horizontal_answers, vertical_answers):
    n = 5  # dimension of the grid
    new_states = []
    
    # Helper function to check if a word can be placed in a specific row
    def can_place_row(word, row, state):
        for col in range(n):
            if state[row][col] is not None and state[row][col] != word[col]:
                return False
        return True

    # Helper function to check if a word can be placed in a specific column
    def can_place_col(word, col, state):
        for row in range(n):
            if state[row][col] is not None and state[row][col] != word[row]:
                return False
        return True

    # Helper function to place a word in a specific row
    def place_row(word, row, state):
        new_state = [list(r) for r in state]  # make a deep copy of the state
        for col in range(n):
            new_state[row][col] = word[col]
        return new_state

    # Helper function to place a word in a specific column
    def place_col(word, col, state):
        new_state = [list(r) for r in state]  # make a deep copy of the state
        for row in range(n):
            new_state[row][col] = word[row]
        return new_state

    # Attempt to place a word in each row
    for i in range(n):
        if all(cell is None for cell in state[i]):  # check if the row is empty
            for word in horizontal_answers[i]:
                if can_place_row(word, i, state):
                    new_states.append(place_row(word, i, state))
    
    # Attempt to place a word in each column
    for j in range(n):
        if all(state[i][j] is None for i in range(n)):  # check if the column is empty
            for word in vertical_answers[j]:
                if can_place_col(word, j, state):
                    new_states.append(place_col(word, j, state))

    return new_states

# Example usage:
initial_state = [
    [None, None, None, None, None],
    [None, None, None, None, None],
    [None, None, None, None, None],
    [None, None, None, None, None],
    [None, None, None, None, None]
]

# Generate the successor states from the initial state
successors = successor(initial_state, horizontal_answers, vertical_answers)
print("Number of successor states:", len(successors))
for s in successors:
    for row in s:
        print(''.join(row if row is not None else 'None'))
    print()

\end{minted}

This function generates successor states by checking each row and column to see if they are completely empty and, if so, attempts to place every possible word from the respective answers list into that row or column. Each time a word is successfully placed (meaning it does not conflict with any existing letters), a new successor state is created. This approach does not yet handle more complex situations where the grid is partially filled and might require backtracking or more sophisticated constraint checking, which would be typical in a full crossword puzzle solver.

\vspace{0.2cm}

{\em User}:
Can you write a goal test as well?

\vspace{0.2cm}

{\em ChatGPT}:
Certainly! The goal test function for a crossword puzzle needs to determine if the crossword is fully filled with valid words according to the given horizontal and vertical clues. This involves verifying that every row and column contains a word from the respective clues list and that these words also agree at every intersection point.

Here's a Python function for the goal test that you can use in conjunction with the successor function:

\begin{minted}{python}
def is_goal_state(state, horizontal_answers, vertical_answers):
    n = 5  # dimension of the grid

    # Helper function to check if a word in a row is valid
    def is_valid_row(word, row_answers):
        return word in row_answers

    # Helper function to check if a word in a column is valid
    def is_valid_column(word, col_answers):
        return word in col_answers

    # Check all rows
    for i in range(n):
        row_word = ''.join(state[i])
        if not is_valid_row(row_word, horizontal_answers[i]):
            return False

    # Check all columns
    for j in range(n):
        col_word = ''.join(state[row][j] for row in range(n))
        if not is_valid_column(col_word, vertical_answers[j]):
            return False

    return True

# Example usage with a test state that is completely filled:
test_state = [
    ['t', 'a', 's', 'k', 's'],
    ['m', 'o', 't', 'o', 'r'],
    ['g', 'r', 'a', 'n', 'd'],
    ['v', 'e', 'n', 'u', 'e'],
    ['j', 'e', 'e', 'r', 's']
]

# Assuming this state matches exactly with one of the rows and columns answers:
\end{minted}

\begin{scriptsize}
\begin{minted}{python}
horizontal_answers = [
    ["tasks", "goals", "plans", "agend", "chores", "works", "deeds", "items", "lists", "brief"],
    ["motor", "power", "drive", "diesel", "steam", "pumps", "crank", "gears", "turbn", "motor"],
    ["grand", "artsy", "showy", "ornate", "fancy", "vain", "proud", "vogue", "swank", "luxus"], 
    ["venue", "salle", "forum", "atria", "lobby", "parls", "court", "malls", "mall", "lobby"], 
    ["jeer", "scoff", "sleer", "deris", "sneer", "scorn", "derid", "gibes", "gibed", "flout"] 
]

vertical_answers = [
    ["tasks", "stack", "hoard", "pile", "store", "heaps", "massy", "gathe", "lumps", "mound"],
    ["motor", "goral", "eland", "lepus", "gazal", "kudu", "oryx", "gnu", "imps", "carb"],
    ["grand", "design", "ettle", "nettle", "sting", "wiles", "plans", "ideas", "plots", "cocks"], 
    ["venue", "nosle", "snout", "mouth", "nostr", "ports", "inlet", "vents", "outlt", "beaks"], 
    ["jeers", "arid", "sere", "parch", "dryer", "wring", "drear", "sear", "pall", "lack"]
]
\end{minted}
\end{scriptsize}

\begin{minted}{python}
print("Is this a goal state?", is_goal_state(test_state, horizontal_answers, vertical_answers))
\end{minted}

This function checks each row and column to ensure that they contain valid words from the provided lists. If all rows and columns contain valid words and the entire grid is filled (no None or unfilled spaces), the crossword is considered complete, and the goal state is achieved.

This example assumes that the state passed to is\_goal\_state is a completed state. For a fully functional solver, you would typically integrate this check into your search or solving algorithm, to be performed after each new state generation.

\vspace{0.2cm}

{\em User}:

Getting this error:
File "./gpt4\_goal4.py", line 15, in is\_goal\_state

\begin{minted}{python}
    row_word = ''.join(state[i])
TypeError: sequence item 0: expected str instance, NoneType found
\end{minted}
\vspace{0.2cm}

{\em ChatGPT}:
The error you're encountering occurs because the function tries to join elements of the list that contains None values, which join() cannot handle because it expects strings. This can happen if the test data provided to the function still contains None values, or if the state isn't fully filled yet.

To address this, you should ensure that the state being tested is fully filled with strings (i.e., no None values). If you want to test partial states or states that may still contain None, the function needs to be adjusted to skip or properly handle these cases.

Here's a revised version of the is\_goal\_state function that checks for None values before attempting to create a string from the row or column:

\begin{minted}{python}
def is_goal_state(state, horizontal_answers, vertical_answers):
    n = 5  # dimension of the grid

    # Helper function to check if a word in a row is valid
    def is_valid_row(word, row_answers):
        return word in row_answers

    # Helper function to check if a word in a column is valid
    def is_valid_column(word, col_answers):
        return word in col_answers

    # Check all rows
    for i in range(n):
        if None in state[i]:  # Check if the row is incomplete
            continue
        row_word = ''.join(state[i])
        if not is_valid_row(row_word, horizontal_answers[i]):
            return False

    # Check all columns
    for j in range(n):
        column = [state[row][j] for row in range(n)]
        if None in column:  # Check if the column is incomplete
            continue
        col_word = ''.join(column)
        if not is_valid_column(col_word, vertical_answers[j]):
            return False

    # Check if the entire grid is filled
    for row in state:
        if None in row:
            return False  # Return False if any part of the grid is not filled

    return True

# Example test
test_state = [
    ['t', 'a', 's', 'k', 's'],
    ['m', 'o', 't', 'o', 'r'],
    ['g', 'r', 'a', 'n', 'd'],
    ['v', 'e', 'n', 'u', 'e'],
    ['j', 'e', 'e', 'r', 's']
]

# Correct assumption for answers:
\end{minted}

\begin{scriptsize}
\begin{minted}{python}
horizontal_answers = [
    ["tasks", "goals", "plans", "agend", "chores", "works", "deeds", "items", "lists", "brief"],
    ["motor", "power", "drive", "diesel", "steam", "pumps", "crank", "gears", "turbn", "motor"],
    ["grand", "artsy", "showy", "ornate", "fancy", "vain", "proud", "vogue", "swank", "luxus"], 
    ["venue", "salle", "forum", "atria", "lobby", "parls", "court", "malls", "mall", "lobby"], 
    ["jeer", "scoff", "sleer", "deris", "sneer", "scorn", "derid", "gibes", "gibed", "flout"] 
]

vertical_answers = [
    ["tasks", "stack", "hoard", "pile", "store", "heaps", "massy", "gathe", "lumps", "mound"],
    ["motor", "goral", "eland", "lepus", "gazal", "kudu", "oryx", "gnu", "imps", "carb"],
    ["grand", "design", "ettle", "nettle", "sting", "wiles", "plans", "ideas", "plots", "cocks"], 
    ["venue", "nosle", "snout", "mouth", "nostr", "ports", "inlet", "vents", "outlt", "beaks"], 
    ["jeers", "arid", "sere", "parch", "dryer", "wring", "drear", "sear", "pall", "lack"]
]
\end{minted}
\end{scriptsize}

\begin{minted}{python}

print("Is this a goal state?", is_goal_state(test_state, horizontal_answers, vertical_answers))
\end{minted}

This revision adds checks to ensure that no None values are present in any row or column before attempting to create a string from them, and it skips validation for incomplete rows or columns. Additionally, it verifies that every part of the grid is filled before confirming it's a goal state.

\vspace{0.2cm}

{\em User}:
I am getting this error, please fix so I don't get an error. If need to please skip over the words that are not exactly 5 character long. 

\vspace{0.2cm}

{\em ChatGPT}:
To address both the error handling and the requirement to skip over words that are not exactly 5 characters long in the grid, I'll adjust the functions to skip these checks and ensure proper handling of grid dimensions and word lengths. Here's an updated version of both the successor and is\_goal\_state functions that includes these conditions:

Revised successor Function:
This function now ensures that only words with exactly 5 characters are considered for placement.

\begin{minted}{python}
def successor(state, horizontal_answers, vertical_answers):
    n = 5  # dimension of the grid
    new_states = []
    
    # Ensure only 5-character words are considered
    horizontal_answers = [[word for word in row if len(word) == 5] for row in horizontal_answers]
    vertical_answers = [[word for word in col if len(word) == 5] for col in vertical_answers]
    
    # Helper function to check if a word can be placed in a specific row
    def can_place_row(word, row, state):
        for col in range(n):
            if state[row][col] is not None and state[row][col] != word[col]:
                return False
        return True

    # Helper function to check if a word can be placed in a specific column
    def can_place_col(word, col, state):
        for row in range(n):
            if state[row][col] is not None and state[row][col] != word[row]:
                return False
        return True

    # Helper function to place a word in a specific row
    def place_row(word, row, state):
        new_state = [list(r) for r in state]  # make a deep copy of the state
        for col in range(n):
            new_state[row][col] = word[col]
        return new_state

    # Helper function to place a word in a specific column
    def place_col(word, col, state):
        new_state = [list(r) for r in state]  # make a deep copy of the state
        for row in range(n):
            new_state[row][col] = word[row]
        return new_state

    # Attempt to place a word in each row
    for i in range(n):
        if all(cell is None for cell in state[i]):  # check if the row is empty
            for word in horizontal_answers[i]:
                if can_place_row(word, i, state):
                    new_states.append(place_row(word, i, state))
    
    # Attempt to place a word in each column
    for j in range(n):
        if all(state[i][j] is None for i in range(n)):  # check if the column is empty
            for word in vertical_answers[j]:
                if can_place_col(word, j, state):
                    new_states.append(place_col(word, j, state))

    return new_states
\end{minted}

Revised is\_goal\_state Function:
This function now ignores words that are not exactly 5 characters long and ensures all cells are filled with valid strings.

\begin{minted}{python}
def is_goal_state(state, horizontal_answers, vertical_answers):
    n = 5  # dimension of the grid

    # Ensure only 5-character words are considered
    horizontal_answers = [[word for word in row if len(word) == 5] for row in horizontal_answers]
    vertical_answers = [[word for word in col if len(word) == 5] for col in vertical_answers]

    # Check all rows
    for i in range(n):
        row_word = ''.join(state[i]) if None not in state[i] else None
        if row_word and row_word not in horizontal_answers[i]:
            return False

    # Check all columns
    for j in range(n):
        col_word = ''.join(state[row][j] for row in range(n)) if None not in [state[row][j] for row in range(n)] else None
        if col_word and col_word not in vertical_answers[j]:
            return False

    return all(None not in row for row in state)  # Ensure entire grid is filled

# Example test state and usage
initial_state = [
    [None, None, None, None, None],
    [None, None, None, None, None],
    [None, None, None, None, None],
    [None, None, None, None, None],
    [None, None, None, None, None]
]

# Generate the successor states from the initial state
successors = successor(initial_state, horizontal_answers, vertical_answers)
print("Number of successor states:", len(successors))
for s in successors:
    for row in s:
        print(''.join(row if row is not None else 'None'))
    print()
\end{minted}

These modifications should handle scenarios where words are not exactly 5 characters long and avoid errors due to incomplete data or mismatches in expected data types.

\subsubsection{Experiment Run 5.}
{\em User}:
The task is a 5x5 mini crosswords. Given an input of possible answers to horizontal clues and vertical clues, come up with a successor state function written in python, where a state is a 2D array representing the current puzzle state (initially all None). The possible clues for each row and each column are given separately. Here is an example:

\begin{scriptsize}
\begin{minted}{python}
horizontal_answers = [
        ["tasks", "goals", "plans", "agend", "chores", "works", "deeds", "items", "lists", "brief"],
        ["motor", "power", "drive", "diesel", "steam", "pumps", "crank", "gears", "turbn", "motor"],
        ["grand", "artsy", "showy", "ornate", "fancy", "vain", "proud", "vogue", "swank", "luxus"], 
        ["venue", "salle", "forum", "atria", "lobby", "parls", "court", "malls", "mall", "lobby"], 
        ["jeer", "scoff", "sleer", "deris", "sneer", "scorn", "derid", "gibes", "gibed", "flout"] 
]

vertical_answers = [
        ["amass", "stack", "hoard", "pile", "store", "heaps", "massy", "gathe", "lumps", "mound"],
        ["nilga", "goral", "eland", "lepus", "gazal", "kudu", "oryx", "gnu", "imps", "carb"],
        ["scheme", "design", "ettle", "nettle", "sting", "wiles", "plans", "ideas", "plots", "cocks"], 
        ["spout", "nosle", "snout", "mouth", "nostr", "ports", "inlet", "vents", "outlt", "beaks"], 
        ["drier", "arid", "sere", "parch", "dryer", "wring", "drear", "sear", "pall", "lack"]
]
\end{minted}
\end{scriptsize}

Note that some of the possible answers are not exactly 5 character long. 
\vspace{0.2cm}

{\em ChatGPT}:
To solve the given 5x5 mini crossword puzzle task, we'll write a Python function that defines the successor state function. This function will generate possible next states based on a given state by trying to fill the crossword grid progressively with valid words from the provided clues, both horizontally and vertically.

Each state in our context is a 2D list representing the crossword grid, where each element can be a letter or a placeholder (like a dot .) indicating an empty space. The successor state function will take the current state and return a list of all valid states that can be reached by making a single valid word placement.

Here’s a simple approach using Python:

\begin{enumerate}
    \item  Start with an empty grid.
    \item For each cell in the grid, if it's empty, try placing each of the potential words that can fit in the row or column.
    \item For each word placement, check if the word is compatible with any existing letters in the intersecting rows or columns.
    \item If a word placement is valid, create a new state with this word placed and return this as a potential successor state.
\end{enumerate}
Let's define the Python function:

\begin{minted}{python}
def successor_state(state, horizontal_answers, vertical_answers):
    import copy
    size = len(state)
    next_states = []

    def is_valid_placement(r, c, word, is_horizontal):
        # Check if the word fits and matches any existing letters.
        for i in range(len(word)):
            if is_horizontal:
                # Check if out of bounds or conflicting letter
                if c + i >= size or (state[r][c + i] is not None and state[r][c + i] != word[i]):
                    return False
            else:
                if r + i >= size or (state[r + i][c] is not None and state[r + i][c] != word[i]):
                    return False
        return True

    def place_word(r, c, word, is_horizontal):
        new_state = copy.deepcopy(state)
        for i in range(len(word)):
            if is_horizontal:
                new_state[r][c + i] = word[i]
            else:
                new_state[r + i][c] = word[i]
        return new_state

    # Try to place each word in every possible starting position
    for r in range(size):
        for c in range(size):
            if state[r][c] is None:
                # Try to place horizontally
                for word in horizontal_answers[r]:
                    if c + len(word) <= size and is_valid_placement(r, c, word, True):
                        next_states.append(place_word(r, c, word, True))
                # Try to place vertically
                for word in vertical_answers[c]:
                    if r + len(word) <= size and is_valid_placement(r, c, word, False):
                        next_states.append(place_word(r, c, word, False))

    return next_states

# Example usage
initial_state = [[None for _ in range(5)] for _ in range(5)]
next_states = successor_state(initial_state, horizontal_answers, vertical_answers)

# Print out the first few successor states to inspect
for state in next_states[:3]:  # Only printing the first 3 for brevity
    for row in state:
        print(''.join([cell if cell is not None else ' ' for cell in row]))
    print()
\end{minted}

This function starts with an empty grid and tries to fill it by placing each possible word in every row and column, respecting the existing letters. The is\_valid\_placement function checks whether a word can be legally placed in a given position, and place\_word updates the grid with the new word. The example usage initializes the grid and finds the next possible states from an empty board.

Keep in mind that this approach is simplistic and might not efficiently solve more complex puzzles due to combinatorial explosion in larger grids or with more word options. For more efficient solving, one might consider using constraint satisfaction techniques or more advanced search algorithms.

\vspace{0.2cm}

{\em User}:
Can you write a goal test as well?

\vspace{0.2cm}

{\em ChatGPT}:

Certainly! A goal test function in the context of a crossword puzzle checks whether the entire grid has been filled with valid words from the given clues without any empty cells remaining. Here's how you could write a goal test function for our crossword puzzle setup, assuming the grid uses None to represent empty cells:

\begin{minted}{python}
def goal_test(state, horizontal_answers, vertical_answers):
    size = len(state)
    
    def word_in_list(word, word_list):
        return word in word_list
    
    def get_row_word(row):
        return ''.join(row)
    
    def get_column_word(state, col_index):
        return ''.join(state[row][col_index] for row in range(size))

    # Check all rows for valid words
    for i in range(size):
        if None in state[i]:  # If there's still a None, it's not complete
            return False
        row_word = get_row_word(state[i])
        if not word_in_list(row_word, horizontal_answers[i]):
            return False

    # Check all columns for valid words
    for j in range(size):
        col_word = get_column_word(state, j)
        if None in col_word:  # Check for None in column
            return False
        if not word_in_list(col_word, vertical_answers[j]):
            return False
    
    return True

# Example usage
# Assuming a complete and valid state (as an example, one must test with actual valid crossword fills)
completed_state = [
    ['t', 'u', 'r', 'b', 'n'],  # Example valid horizontal word
    ['s', 'c', 'h', 'e', 'm'],  # Example valid horizontal word
    ['a', 't', 'i', 'n', 'g'],  # Example valid horizontal word
    ['l', 'e', 'p', 'u', 's'],  # Example valid horizontal word
    ['l', 'o', 'b', 'b', 'y']   # Example valid horizontal word
]
print("Is the crossword solved? ", goal_test(completed_state, horizontal_answers, vertical_answers))
\end{minted}

Here’s what this goal\_test function does:

\begin{enumerate}
    \item  Check Rows and Columns: The function loops through each row and column of the state.
    \item  Validation of Content: For each row and column, it forms the word by joining the characters. If any cell in the row or column is None, it means the crossword is incomplete, so it returns False.
    \item  Word Verification: It checks if the formed word is in the list of possible answers for that row or column. If not, the function returns False.
    \item  Completion Check: If all rows and columns have valid words and no None values, it returns True.
\end{enumerate}
This setup ensures that your goal test is strictly validating both the completion of the grid and the correctness according to the provided answers. Make sure that the example state you test with corresponds to possible valid completions as per your horizontal\_answers and vertical\_answers.

\vspace{0.2cm}

{\em User}:
I am getting this error "line 25, in goal\_test

\begin{minted}{python}
    if None in col_word:  # Check for None in column
TypeError: 'in <string>' requires string as left operand, not NoneType"
\end{minted}

\vspace{0.2cm}

{\em ChatGPT}:

The error you're encountering arises because the function get\_column\_word joins elements into a string, and if any of those elements are None, the process of joining fails to include them directly. However, the immediate issue seems to be the misunderstanding in the error message. When converting None types to a string during the joining process, it doesn't handle None explicitly, thus trying to check None in the resulting string is not directly possible as None would not exist in a string.

To correct the function and the error, you need to ensure that None checks are performed before attempting to join the elements. Let's adjust the get\_column\_word and its usage in the goal\_test function to properly handle columns with None:

\begin{minted}{python}
def goal_test(state, horizontal_answers, vertical_answers):
    size = len(state)
    
    def word_in_list(word, word_list):
        return word in word_list
    
    def get_row_word(row):
        # Join only if no None in row
        if None in row:
            return None
        return ''.join(row)
    
    def get_column_word(state, col_index):
        column = [state[row][col_index] for row in range(size)]
        # Join only if no None in column
        if None in column:
            return None
        return ''.join(column)

    # Check all rows for valid words
    for i in range(size):
        row_word = get_row_word(state[i])
        if row_word is None or not word_in_list(row_word, horizontal_answers[i]):
            return False

    # Check all columns for valid words
    for j in range(size):
        col_word = get_column_word(state, j)
        if col_word is None or not word_in_list(col_word, vertical_answers[j]):
            return False
    
    return True

# Example usage
# Assuming a complete and valid state (as an example, one must test with actual valid crossword fills)
completed_state = [
    ['t', 'u', 'r', 'b', 'n'],  # Example valid horizontal word
    ['s', 'c', 'h', 'e', 'm'],  # Example valid horizontal word
    ['a', 't', 'i', 'n', 'g'],  # Example valid horizontal word
    ['l', 'e', 'p', 'u', 's'],  # Example valid horizontal word
    ['l', 'o', 'b', 'b', 'y']   # Example valid horizontal word
]
print("Is the crossword solved? ", goal_test(completed_state, horizontal_answers, vertical_answers))
\end{minted}

Changes Made:

\begin{enumerate}
    \item Both get\_row\_word and get\_column\_word functions now check for None before attempting to join elements into a word. If None is present, these functions return None.
    \item This way, when you check row\_word and col\_word in the goal\_test function, you simply test if they are None or not valid according to the clues lists. This avoids trying to work with invalid types and addresses the error you were encountering.
\end{enumerate}
This should resolve the error and improve the robustness of the goal\_test function.

\subsection{BlocksWorld}

The prompt for the successor function is as follows.\\

{\em 
I am playing with a set of blocks where I need to arrange the blocks into stacks. Here are the actions I can do

   Pick up a block
   Unstack a block from on top of another block
   Put down a block
   Stack a block on top of another block

   I have the following restrictions on my actions:
   I can only pick up or unstack one block at a time.
   I can only pick up or unstack a block if my hand is empty.
   I can only pick up a block if the block is on the table and the block is clear. A block is clear if the block has no other blocks on top of it and if the block is not picked up.
   I can only unstack a block from on top of another block if the block I am unstacking was really on top of the other block.
   I can only unstack a block from on top of another block if the block I am unstacking is clear.
   Once I pick up or unstack a block, I am holding the block.
   I can only put down a block that I am holding.
   I can only stack a block on top of another block if I am holding the block being stacked.
   I can only stack a block on top of another block if the block onto which I am stacking the block is clear.
   Once I put down or stack a block, my hand becomes empty.

If a state is defined by a list of following predicates "clear block", "on-table block", "arm-empty", "holding block", and "on block1 block2", how are successor states defined? Provide only the python code that returns a list of successor states for a state. 
}

The prompt for the goal test is as follows.\\
{\em 
Provide only the python code that tests whether a state is a goal state. Assume that the goal configuration is provided in the same format.
}

\subsubsection{Experiment Run 1.}
{\em User}:
I am playing with a set of blocks where I need to arrange the blocks into stacks. Here are the actions I can do

   Pick up a block
   Unstack a block from on top of another block
   Put down a block
   Stack a block on top of another block

   I have the following restrictions on my actions:
   I can only pick up or unstack one block at a time.
   I can only pick up or unstack a block if my hand is empty.
   I can only pick up a block if the block is on the table and the block is clear. A block is clear if the block has no other blocks on top of it and if the block is not picked up.
   I can only unstack a block from on top of another block if the block I am unstacking was really on top of the other block.
   I can only unstack a block from on top of another block if the block I am unstacking is clear.
   Once I pick up or unstack a block, I am holding the block.
   I can only put down a block that I am holding.
   I can only stack a block on top of another block if I am holding the block being stacked.
   I can only stack a block on top of another block if the block onto which I am stacking the block is clear.
   Once I put down or stack a block, my hand becomes empty.

If a state is defined by a list of following predicates "clear block", "on-table block", "arm-empty", "holding block", and "on block1 block2", how are successor states defined? Provide only the python code that returns a list of successor states for a state. 

\vspace{0.2cm}

{\em ChatGPT}:

To solve this problem, we can write Python code that represents each state of the blocks world and generates all valid successor states based on the actions that can be taken. The state can be represented by a dictionary where each predicate (like "clear", "on-table", "holding", "on") is a key with a list of items (blocks or tuples of blocks) that satisfy the predicate. Here's a Python function to generate successor states:

\begin{minted}{python}
def successor_states(state):
    successors = []
    import copy

    # Unpack the current state
    clear_blocks = state.get("clear", [])
    on_table_blocks = state.get("on-table", [])
    arm_empty = state["arm-empty"]
    holding_block = state.get("holding", None)
    on_blocks = state.get("on", [])

    if arm_empty:
        # Try to pick up clear blocks from the table
        for block in clear_blocks:
            if block in on_table_blocks:
                new_state = copy.deepcopy(state)
                new_state["holding"] = block
                new_state["arm-empty"] = False
                new_state["on-table"].remove(block)
                new_state["clear"].remove(block)
                successors.append(new_state)

        # Try to unstack any clear block from another block
        for (top, bottom) in on_blocks:
            if top in clear_blocks:
                new_state = copy.deepcopy(state)
                new_state["holding"] = top
                new_state["arm-empty"] = False
                new_state["on"].remove((top, bottom))
                new_state["clear"].append(bottom)
                new_state["clear"].remove(top)
                successors.append(new_state)
    else:
        # Current state involves holding a block
        block_held = holding_block
        
        # Try to put the held block on the table
        new_state = copy.deepcopy(state)
        new_state["on-table"].append(block_held)
        new_state["clear"].append(block_held)
        new_state["arm-empty"] = True
        new_state["holding"] = None
        successors.append(new_state)

        # Try to stack the held block on a clear block on the table
        for block in clear_blocks:
            if block != block_held:  # Cannot stack on itself
                new_state = copy.deepcopy(state)
                new_state["on"].append((block_held, block))
                new_state["clear"].remove(block)
                new_state["clear"].append(block_held)
                new_state["arm-empty"] = True
                new_state["holding"] = None
                successors.append(new_state)

    return successors
\end{minted}
This function performs the following:

When the arm is empty, it attempts to pick up clear blocks from the table or unstack blocks.
When a block is being held, it considers putting the block down on the table or stacking it onto another clear block.
Each action generates a new state which is added to the list of successors.
The code assumes that the initial state is provided with the correct format and content, and does not handle any kind of errors or inconsistencies in the input state. You can test this function with different states to explore how it generates successors based on the actions allowed.

\vspace{0.2cm}

{\em User}:
Provide only the python code that tests whether a state is a goal state. Assume that the goal configuration is provided in the same dictionary format. 

\vspace{0.2cm}

{\em ChatGPT}:

To determine if a given state matches a goal configuration in the blocks world, you need to compare relevant predicates that define the configuration of blocks. The goal might involve checking specific blocks being "on" other blocks, blocks being on the table, and the state of being "clear". Here's a Python function to check if a state is a goal state:

\begin{minted}{python}
def is_goal_state(state, goal_state):
    # Check if all required "on" relationships in the goal are met
    if "on" in goal_state:
        goal_on = set(tuple(pair) for pair in goal_state["on"])
        current_on = set(tuple(pair) for pair in state.get("on", []))
        if not goal_on.issubset(current_on):
            return False

    # Check if all required "on-table" blocks in the goal are met
    if "on-table" in goal_state:
        goal_on_table = set(goal_state["on-table"])
        current_on_table = set(state.get("on-table", []))
        if not goal_on_table.issubset(current_on_table):
            return False

    # Check if all blocks that need to be clear in the goal are clear
    if "clear" in goal_state:
        goal_clear = set(goal_state["clear"])
        current_clear = set(state.get("clear", []))
        if not goal_clear.issubset(current_clear):
            return False

    return True
\end{minted}

This function works as follows:

It checks that all on relationships defined in the goal state are present in the current state.
It checks that all blocks that should be on the table according to the goal state are indeed on the table in the current state.
It checks that all blocks that should be clear according to the goal state are clear in the current state.
The function returns True if all conditions are met, meaning the current state matches the goal state, and False otherwise. This function assumes that the state and goal\_state dictionaries are properly formatted and contain the necessary keys. Adjustments might be needed based on the exact structure of your state definitions.

\vspace{0.2cm}

{\em User}:
The successor\_states function uses a shallow copy, please rewrite with deep copy

\vspace{0.2cm}

{\em ChatGPT}:
To ensure that the changes to nested lists or dictionaries in the state don't affect other states, it's necessary to use a deep copy of the state in the successor\_states function. Here is the revised code using Python's copy module for deep copying:

\begin{minted}{python}
import copy

def successor_states(state):
    successors = []

    # Unpack the current state
    clear_blocks = state.get("clear", [])
    on_table_blocks = state.get("on-table", [])
    arm_empty = state["arm-empty"]
    holding_block = state.get("holding", None)
    on_blocks = state.get("on", [])

    if arm_empty:
        # Try to pick up clear blocks from the table
        for block in clear_blocks:
            if block in on_table_blocks:
                new_state = copy.deepcopy(state)
                new_state["holding"] = block
                new_state["arm-empty"] = False
                new_state["on-table"].remove(block)
                new_state["clear"].remove(block)
                successors.append(new_state)

        # Try to unstack any clear block from another block
        for (top, bottom) in on_blocks:
            if top in clear_blocks:
                new_state = copy.deepcopy(state)
                new_state["holding"] = top
                new_state["arm-empty"] = False
                new_state["on"].remove((top, bottom))
                new_state["clear"].append(bottom)
                new_state["clear"].remove(top)
                successors.append(new_state)
    else:
        # Current state involves holding a block
        block_held = holding_block
        
        # Try to put the held block on the table
        new_state = copy.deepcopy(state)
        new_state["on-table"].append(block_held)
        new_state["clear"].append(block_held)
        new_state["arm-empty"] = True
        new_state["holding"] = None
        successors.append(new_state)

        # Try to stack the held block on a clear block on the table
        for block in clear_blocks:
            if block != block_held:  # Cannot stack on itself
                new_state = copy.deepcopy(state)
                new_state["on"].append((block_held, block))
                new_state["clear"].remove(block)
                new_state["clear"].append(block_held)
                new_state["arm-empty"] = True
                new_state["holding"] = None
                successors.append(new_state)

    return successors
\end{minted}

In this code, copy.deepcopy(state) is used to create a new independent copy of the state dictionary. This prevents modifications to the new\_state from affecting the original state object or other generated successor states. This deep copy is essential when dealing with mutable objects like lists or dictionaries within the state that may change from action to action.

\subsubsection{Experiment Run 2.}
{\em User}:
I am playing with a set of blocks where I need to arrange the blocks into stacks. Here are the actions I can do

   Pick up a block
   Unstack a block from on top of another block
   Put down a block
   Stack a block on top of another block

   I have the following restrictions on my actions:
   I can only pick up or unstack one block at a time.
   I can only pick up or unstack a block if my hand is empty.
   I can only pick up a block if the block is on the table and the block is clear. A block is clear if the block has no other blocks on top of it and if the block is not picked up.
   I can only unstack a block from on top of another block if the block I am unstacking was really on top of the other block.
   I can only unstack a block from on top of another block if the block I am unstacking is clear.
   Once I pick up or unstack a block, I am holding the block.
   I can only put down a block that I am holding.
   I can only stack a block on top of another block if I am holding the block being stacked.
   I can only stack a block on top of another block if the block onto which I am stacking the block is clear.
   Once I put down or stack a block, my hand becomes empty.

If a state is defined by a list of following predicates "clear block", "on-table block", "arm-empty", "holding block", and "on block1 block2", how are successor states defined? Provide only the python code that returns a list of successor states for a state. 

\vspace{0.2cm}

{\em ChatGPT}:

To define a function that calculates the successor states based on the rules you provided, we'll write a Python function that takes the current state as input. The state is represented by a list of predicates such as "clear blockA", "on-table blockB", "arm-empty", "holding blockC", and "on blockX blockY".

Here’s a breakdown of how each action affects the state:

Pick up a block - This action can be taken if:

The block is on the table.
The block is clear.
The arm is empty.
It results in the block being held by the arm and the arm no longer being empty.
Unstack a block - This action can be taken if:

The block is on top of another block.
The block is clear.
The arm is empty.
It results in the block being held, the block below it becoming clear, and the arm no longer being empty.
Put down a block - This action can be taken if:

The block is being held.
It results in the block being on the table, clear, and the arm becoming empty.
Stack a block - This action can be taken if:

A block is being held.
The target block is clear.
It results in the block being on top of the target block, the target block no longer clear, and the arm becoming empty.
Let's write a Python function that generates successor states for a given state:

\begin{minted}{python}
def generate_successor_states(state):
    successors = []
    clear_blocks = {pred.split()[1] for pred in state if pred.startswith("clear")}
    on_table_blocks = {pred.split()[1] for pred in state if pred.startswith("on-table")}
    holding_block = next((pred.split()[1] for pred in state if pred.startswith("holding")), None)
    arm_empty = "arm-empty" in state
    on_top = {(pred.split()[1], pred.split()[2]) for pred in state if pred.startswith("on")}

    # Pick up a block
    if arm_empty:
        for block in clear_blocks & on_table_blocks:
            new_state = state[:]
            new_state.remove(f"clear {block}")
            new_state.remove(f"on-table {block}")
            new_state.remove("arm-empty")
            new_state.append(f"holding {block}")
            successors.append(new_state)

    # Unstack a block
    if arm_empty:
        for top, bottom in on_top:
            if top in clear_blocks:
                new_state = state[:]
                new_state.remove(f"clear {top}")
                new_state.remove(f"on {top} {bottom}")
                new_state.remove("arm-empty")
                new_state.append(f"holding {top}")
                new_state.append(f"clear {bottom}")
                successors.append(new_state)

    # Put down a block
    if holding_block:
        new_state = state[:]
        new_state.remove(f"holding {holding_block}")
        new_state.append(f"on-table {holding_block}")
        new_state.append(f"clear {holding_block}")
        new_state.append("arm-empty")
        successors.append(new_state)

    # Stack a block
    if holding_block:
        for block in clear_blocks:
            if block != holding_block:
                new_state = state[:]
                new_state.remove(f"holding {holding_block}")
                new_state.remove(f"clear {block}")
                new_state.append(f"on {holding_block} {block}")
                new_state.append("arm-empty")
                successors.append(new_state)

    return successors
\end{minted}

This function computes the successor states by iterating through possible actions and checking the constraints of each action. It then returns a list of new states based on the applicable actions.

\vspace{0.2cm}

{\em User}:
Provide only the python code that tests whether a state is a goal state. Assume that the goal configuration is provided in the same format.

\vspace{0.2cm}

{\em ChatGPT}:
To check whether a given state matches a goal configuration, you can write a Python function that compares the two states. Here's a simple Python function that does this by ensuring every predicate in the goal state is present in the current state:

\begin{minted}{python}
def is_goal_state(current_state, goal_state):
    current_state_set = set(current_state)
    goal_state_set = set(goal_state)
    return goal_state_set <= current_state_set
\end{minted}

This function works by converting both the current state and the goal state into sets and then checking if all elements of the goal state set are contained within the current state set. This approach handles cases where the current state might have extra predicates not relevant to the goal, as long as all predicates describing the goal configuration are satisfied.

\vspace{0.2cm}

{\em User}:
The generate\_successor\_states function in the case of stack a block does not add clear for the block that it was holding. Please rewrite.

\vspace{0.2cm}

{\em ChatGPT}:
You're correct about the issue in the "Stack a block" section of the generate\_successor\_states function where it does not correctly handle the clearing status of the block being held after it is stacked. Here’s the corrected version of the function that ensures the state is correctly updated:

\begin{minted}{python}
def generate_successor_states(state):
    successors = []
    clear_blocks = {pred.split()[1] for pred in state if pred.startswith("clear")}
    on_table_blocks = {pred.split()[1] for pred in state if pred.startswith("on-table")}
    holding_block = next((pred.split()[1] for pred in state if pred.startswith("holding")), None)
    arm_empty = "arm-empty" in state
    on_top = {(pred.split()[1], pred.split()[2]) for pred in state if pred.startswith("on")}

    # Pick up a block
    if arm_empty:
        for block in clear_blocks & on_table_blocks:
            new_state = state[:]
            new_state.remove(f"clear {block}")
            new_state.remove(f"on-table {block}")
            new_state.remove("arm-empty")
            new_state.append(f"holding {block}")
            successors.append(new_state)

    # Unstack a block
    if arm_empty:
        for top, bottom in on_top:
            if top in clear_blocks:
                new_state = state[:]
                new_state.remove(f"clear {top}")
                new_state.remove(f"on {top} {bottom}")
                new_state.remove("arm-empty")
                new_state.append(f"holding {top}")
                new_state.append(f"clear {bottom}")
                successors.append(new_state)

    # Put down a block
    if holding_block:
        new_state = state[:]
        new_state.remove(f"holding {holding_block}")
        new_state.append(f"on-table {holding_block}")
        new_state.append(f"clear {holding_block}")
        new_state.append("arm-empty")
        successors.append(new_state)

    # Stack a block
    if holding_block:
        for block in clear_blocks:
            if block != holding_block:
                new_state = state[:]
                new_state.remove(f"holding {holding_block}")
                new_state.remove(f"clear {block}")
                new_state.append(f"on {holding_block} {block}")
                new_state.append(f"clear {holding_block}")
                new_state.append("arm-empty")
                successors.append(new_state)

    return successors
\end{minted}

Now, when stacking a block, the function correctly adds the clear predicate for the block that was previously held, reflecting the fact that after being placed on another block, it is accessible again (unless it gets covered later). This is important for maintaining an accurate description of the state for future actions.

\vspace{0.2cm}

{\em User}:
Also, when getting the on\_top blocks from the state, we must differentiate the predicate "on x y" from "on-table x". They both start with "on". Please fix.

\vspace{0.2cm}

{\em ChatGPT}:
You're right; the current way the function checks for blocks on top of each other can mistakenly include "on-table" predicates due to the substring match. To differentiate properly between "on x y" and "on-table x", we can use a more precise filtering method. Here’s the updated version of the generate\_successor\_states function with this correction:

\begin{minted}{python}
def generate_successor_states(state):
    successors = []
    clear_blocks = {pred.split()[1] for pred in state if pred.startswith("clear")}
    on_table_blocks = {pred.split()[1] for pred in state if pred.startswith("on-table")}
    holding_block = next((pred.split()[1] for pred in state if pred.startswith("holding")), None)
    arm_empty = "arm-empty" in state
    on_top = {(pred.split()[1], pred.split()[2]) for pred in state if pred.startswith("on ") and len(pred.split()) == 3}

    # Pick up a block
    if arm_empty:
        for block in clear_blocks & on_table_blocks:
            new_state = state[:]
            new_state.remove(f"clear {block}")
            new_state.remove(f"on-table {block}")
            new_state.remove("arm-empty")
            new_state.append(f"holding {block}")
            successors.append(new_state)

    # Unstack a block
    if arm_empty:
        for top, bottom in on_top:
            if top in clear_blocks:
                new_state = state[:]
                new_state.remove(f"clear {top}")
                new_state.remove(f"on {top} {bottom}")
                new_state.remove("arm-empty")
                new_state.append(f"holding {top}")
                new_state.append(f"clear {bottom}")
                successors.append(new_state)

    # Put down a block
    if holding_block:
        new_state = state[:]
        new_state.remove(f"holding {holding_block}")
        new_state.append(f"on-table {holding_block}")
        new_state.append(f"clear {holding_block}")
        new_state.append("arm-empty")
        successors.append(new_state)

    # Stack a block
    if holding_block:
        for block in clear_blocks:
            if block != holding_block:
                new_state = state[:]
                new_state.remove(f"holding {holding_block}")
                new_state.remove(f"clear {block}")
                new_state.append(f"on {holding_block} {block}")
                new_state.append(f"clear {holding_block}")
                new_state.append("arm-empty")
                successors.append(new_state)

    return successors
\end{minted}

This update ensures that the function specifically checks for predicates that exactly match the format "on x y" by splitting the string into exactly three parts, which helps accurately identify the "on" relationships between blocks, without confusing them with "on-table" predicates.

\subsubsection{Experiment Run 3.}


{\em User}:
I am playing with a set of blocks where I need to arrange the blocks into stacks. Here are the actions I can do

   Pick up a block
   Unstack a block from on top of another block
   Put down a block
   Stack a block on top of another block

   I have the following restrictions on my actions:
   I can only pick up or unstack one block at a time.
   I can only pick up or unstack a block if my hand is empty.
   I can only pick up a block if the block is on the table and the block is clear. A block is clear if the block has no other blocks on top of it and if the block is not picked up.
   I can only unstack a block from on top of another block if the block I am unstacking was really on top of the other block.
   I can only unstack a block from on top of another block if the block I am unstacking is clear.
   Once I pick up or unstack a block, I am holding the block.
   I can only put down a block that I am holding.
   I can only stack a block on top of another block if I am holding the block being stacked.
   I can only stack a block on top of another block if the block onto which I am stacking the block is clear.
   Once I put down or stack a block, my hand becomes empty.

If a state is defined by a list of following predicates "clear block", "on-table block", "arm-empty", "holding block", and "on block1 block2", how are successor states defined? Provide only the python code that returns a list of successor states for a state.

\vspace{0.2cm}

{\em ChatGPT}:
To define the successor states for a block stacking game based on the rules and actions you've described, we can create a Python function that checks the current state against possible actions (Pick up, Put down, Stack, Unstack) and returns all valid successor states. Each state is a set of predicates that describe the current situation, such as which blocks are clear, which are on the table, whether the arm is empty, which block is being held, and the stacking relations between blocks.

Here’s a Python function to calculate successor states:

\begin{minted}{python}
def get_successor_states(state):
    successor_states = []
    arm_empty = 'arm-empty' in state
    held_block = None
    on_table_blocks = set()
    clear_blocks = set()
    stack_relations = {}

    # Parsing the current state
    for predicate in state:
        if predicate.startswith('on-table'):
            _, block = predicate.split()
            on_table_blocks.add(block)
        elif predicate.startswith('clear'):
            _, block = predicate.split()
            clear_blocks.add(block)
        elif predicate.startswith('holding'):
            _, block = predicate.split()
            held_block = block
        elif predicate.startswith('on'):
            _, top, bottom = predicate.split()
            stack_relations[top] = bottom

    # Possible actions if the arm is empty
    if arm_empty:
        # Option to pick up a block from the table
        for block in clear_blocks.intersection(on_table_blocks):
            new_state = state.copy()
            new_state.remove('arm-empty')
            new_state.remove(f'clear {block}')
            new_state.remove(f'on-table {block}')
            new_state.add(f'holding {block}')
            successor_states.append(new_state)

        # Option to unstack one block from another
        for top, bottom in stack_relations.items():
            if top in clear_blocks:
                new_state = state.copy()
                new_state.remove('arm-empty')
                new_state.remove(f'clear {top}')
                new_state.remove(f'on {top} {bottom}')
                new_state.add(f'holding {top}')
                new_state.add(f'clear {bottom}')
                successor_states.append(new_state)

    # Possible actions if holding a block
    elif held_block:
        # Option to put down a block on the table
        new_state = state.copy()
        new_state.add('arm-empty')
        new_state.add(f'clear {held_block}')
        new_state.add(f'on-table {held_block}')
        new_state.remove(f'holding {held_block}')
        successor_states.append(new_state)

        # Option to stack the held block on a clear block
        for block in clear_blocks:
            if block != held_block:  # Can't stack on itself
                new_state = state.copy()
                new_state.add('arm-empty')
                new_state.add(f'on {held_block} {block}')
                new_state.remove(f'clear {block}')
                new_state.remove(f'holding {held_block}')
                successor_states.append(new_state)

    return successor_states
\end{minted}

This function performs the following steps:

Parse the current state to identify clear blocks, blocks on the table, the block being held, and relationships between stacked blocks.
Determine possible actions based on whether the arm is empty or holding a block.
Generate successor states for each valid action, taking care to update predicates accurately to reflect the new state of the world after the action.
Return all successor states.
You can use this function by providing it with a set of predicates that describe a current state, and it will return a list of all possible successor states.

\vspace{0.2cm}

{\em User}:
The generate\_successor\_states function in the case of stack a block does not add clear for the block that it was holding. Please rewrite. 

\vspace{0.2cm}

{\em ChatGPT}:
You're right; when stacking a block, the previously held block should still be clear after the action, since nothing is on top of it in the new state. I'll update the code to handle this correctly. Here's the revised version of the function:

\begin{minted}{python}
def get_successor_states(state):
    successor_states = []
    arm_empty = 'arm-empty' in state
    held_block = None
    on_table_blocks = set()
    clear_blocks = set()
    stack_relations = {}

    # Parsing the current state
    for predicate in state:
        if predicate.startswith('on-table'):
            _, block = predicate.split()
            on_table_blocks.add(block)
        elif predicate.startswith('clear'):
            _, block = predicate.split()
            clear_blocks.add(block)
        elif predicate.startswith('holding'):
            _, block = predicate.split()
            held_block = block
        elif predicate.startswith('on'):
            _, top, bottom = predicate.split()
            stack_relations[top] = bottom

    # Possible actions if the arm is empty
    if arm_empty:
        # Option to pick up a block from the table
        for block in clear_blocks.intersection(on_table_blocks):
            new_state = state.copy()
            new_state.remove('arm-empty')
            new_state.remove(f'clear {block}')
            new_state.remove(f'on-table {block}')
            new_state.add(f'holding {block}')
            successor_states.append(new_state)

        # Option to unstack one block from another
        for top, bottom in stack_relations.items():
            if top in clear_blocks:
                new_state = state.copy()
                new_state.remove('arm-empty')
                new_state.remove(f'clear {top}')
                new_state.remove(f'on {top} {bottom}')
                new_state.add(f'holding {top}')
                new_state.add(f'clear {bottom}')
                successor_states.append(new_state)

    # Possible actions if holding a block
    elif held_block:
        # Option to put down a block on the table
        new_state = state.copy()
        new_state.add('arm-empty')
        new_state.add(f'clear {held_block}')
        new_state.add(f'on-table {held_block}')
        new_state.remove(f'holding {held_block}')
        successor_states.append(new_state)

        # Option to stack the held block on a clear block
        for block in clear_blocks:
            if block != held_block:  # Can't stack on itself
                new_state = state.copy()
                new_state.add('arm-empty')
                new_state.add(f'on {held_block} {block}')
                new_state.add(f'clear {held_block}')  # Now the held block is clear
                new_state.remove(f'clear {block}')
                new_state.remove(f'holding {held_block}')
                successor_states.append(new_state)

    return successor_states
\end{minted}

In this corrected version, when the held block is stacked on another block, the clear predicate for the held block is now added to the state to indicate that nothing is on top of it. This maintains the integrity of the game's logic and ensures the state reflects the actual conditions after each move.

\vspace{0.2cm}

{\em User}:

Provide only the python code that tests whether a state is a goal state. Assume that the goal configuration is provided in the same format.
\vspace{0.2cm}

{\em ChatGPT}:
Here's a Python function that checks whether a given state matches a specified goal state. This function assumes both the current state and the goal state are provided as sets of predicates in the same format. The function will return True if all predicates in the goal state are present in the current state, indicating that the current state meets all the conditions of the goal state.

\begin{minted}{python}
def is_goal_state(current_state, goal_state):
    """
    Check if the current state matches the goal state.
    
    Args:
    current_state (set): A set of predicates describing the current state.
    goal_state (set): A set of predicates describing the goal state.
    
    Returns:
    bool: True if the current state matches the goal state, False otherwise.
    """
    return goal_state.issubset(current_state)
\end{minted}

This function uses the issubset method of Python sets, which returns True if all elements of the goal\_state are present in the current\_state. This approach ensures that the current state contains at least all the conditions defined by the goal, but it may also contain additional predicates not specified in the goal, reflecting the possibility of multiple configurations that satisfy the goal requirements.


\subsubsection{Experiment Run 4.}
{\em User}:
I am playing with a set of blocks where I need to arrange the blocks into stacks. Here are the actions I can do

   Pick up a block
   Unstack a block from on top of another block
   Put down a block
   Stack a block on top of another block

   I have the following restrictions on my actions:
   I can only pick up or unstack one block at a time.
   I can only pick up or unstack a block if my hand is empty.
   I can only pick up a block if the block is on the table and the block is clear. A block is clear if the block has no other blocks on top of it and if the block is not picked up.
   I can only unstack a block from on top of another block if the block I am unstacking was really on top of the other block.
   I can only unstack a block from on top of another block if the block I am unstacking is clear.
   Once I pick up or unstack a block, I am holding the block.
   I can only put down a block that I am holding.
   I can only stack a block on top of another block if I am holding the block being stacked.
   I can only stack a block on top of another block if the block onto which I am stacking the block is clear.
   Once I put down or stack a block, my hand becomes empty.

If a state is defined by a list of following predicates "clear block", "on-table block", "arm-empty", "holding block", and "on block1 block2", how are successor states defined? Provide only the python code that returns a list of successor states for a state. 
\vspace{0.2cm}

{\em ChatGPT}:
To define the successor states for a given state in this blocks world, we can implement a function in Python. This function will take the current state as input and return a list of all possible successor states based on the actions described (pick up, put down, stack, unstack). Each state can be represented as a set of predicates that describe the conditions in the environment.

Here is the Python code that defines the function to generate successor states:

\begin{minted}{python}
def successor_states(state):
    successors = []

    # Helpers to check conditions and update states
    def is_clear(block):
        return f"clear {block}" in state

    def is_on_table(block):
        return f"on-table {block}" in state

    def is_holding(block):
        return f"holding {block}" in state

    def is_arm_empty():
        return "arm-empty" in state

    def find_block_on_top(block):
        for predicate in state:
            if predicate.startswith("on ") and predicate.endswith(block):
                return predicate.split()[1]
        return None

    # Actions
    def pick_up(block):
        if is_on_table(block) and is_clear(block) and is_arm_empty():
            new_state = state.copy()
            new_state.remove(f"on-table {block}")
            new_state.remove(f"clear {block}")
            new_state.remove("arm-empty")
            new_state.add(f"holding {block}")
            successors.append(new_state)

    def put_down(block):
        if is_holding(block):
            new_state = state.copy()
            new_state.add(f"on-table {block}")
            new_state.add(f"clear {block}")
            new_state.add("arm-empty")
            new_state.remove(f"holding {block}")
            successors.append(new_state)

    def unstack(block, below):
        if is_clear(block) and f"on {block} {below}" in state and is_arm_empty():
            new_state = state.copy()
            new_state.remove(f"on {block} {below}")
            new_state.add(f"clear {below}")
            new_state.remove("arm-empty")
            new_state.remove(f"clear {block}")
            new_state.add(f"holding {block}")
            successors.append(new_state)

    def stack(block, target):
        if is_holding(block) and is_clear(target):
            new_state = state.copy()
            new_state.add(f"on {block} {target}")
            new_state.remove(f"clear {target}")
            new_state.add("arm-empty")
            new_state.remove(f"holding {block}")
            successors.append(new_state)

    # Try each action with each block
    blocks = {predicate.split()[-1] for predicate in state if "block" in predicate.split()[-1]}
    for block in blocks:
        pick_up(block)
        put_down(block)
        for other_block in blocks:
            if block != other_block:
                unstack(block, other_block)
                stack(block, other_block)

    return successors

# Example usage:
current_state = set(["clear A", "on-table A", "clear B", "on-table B", "arm-empty"])
successor_states(current_state)
\end{minted}

This code defines four actions (pick up, put down, unstack, stack) and checks for the necessary conditions before applying an action. Each action, when applicable, modifies the state accordingly and adds the new state to the list of successors. The function assumes that each state is uniquely represented by a set of predicates and each block involved in the predicates is uniquely identifiable.

\vspace{0.2cm}

{\em User}:
Provide only the python code that tests whether a state is a goal state. Assume that the goal configuration is provided in the same format.
\vspace{0.2cm}

{\em ChatGPT}:
To test whether a given state is a goal state in the blocks world, you can implement a Python function that compares the current state against a specified goal configuration. The function will check if all the predicates in the goal configuration are present in the current state. Here's the Python code for this function:

\begin{minted}{python}
def is_goal_state(current_state, goal_state):
    """
    Checks if the current state meets all the conditions of the goal state.

    Parameters:
    current_state (set): A set of predicates representing the current state.
    goal_state (set): A set of predicates representing the goal configuration.

    Returns:
    bool: True if the current state is a goal state, False otherwise.
    """
    return goal_state <= current_state
\end{minted}

This function uses a set containment check (<=), which checks if all elements of the goal state are present in the current state, thus ensuring that all conditions required by the goal are satisfied in the current state.

\vspace{0.2cm}

{\em User}:
The successor\_states function in the case of stack a block does not add clear for the block that it was holding. Please rewrite. 
\vspace{0.2cm}

{\em ChatGPT}:
You're correct; in the case of stacking a block on top of another, the code should ensure that the block being stacked becomes "clear" after the action, as there would be nothing on top of it. Let's revise the stack function within the successor\_states method to correctly reflect this:

\begin{minted}{python}
def successor_states(state):
    successors = []

    # Helpers to check conditions and update states
    def is_clear(block):
        return f"clear {block}" in state

    def is_on_table(block):
        return f"on-table {block}" in state

    def is_holding(block):
        return f"holding {block}" in state

    def is_arm_empty():
        return "arm-empty" in state

    def find_block_on_top(block):
        for predicate in state:
            if predicate.startswith("on ") and predicate.endswith(block):
                return predicate.split()[1]
        return None

    # Actions
    def pick_up(block):
        if is_on_table(block) and is_clear(block) and is_arm_empty():
            new_state = state.copy()
            new_state.remove(f"on-table {block}")
            new_state.remove(f"clear {block}")
            new_state.remove("arm-empty")
            new_state.add(f"holding {block}")
            successors.append(new_state)

    def put_down(block):
        if is_holding(block):
            new_state = state.copy()
            new_state.add(f"on-table {block}")
            new_state.add(f"clear {block}")
            new_state.add("arm-empty")
            new_state.remove(f"holding {block}")
            successors.append(new_state)

    def unstack(block, below):
        if is_clear(block) and f"on {block} {below}" in state and is_arm_empty():
            new_state = state.copy()
            new_state.remove(f"on {block} {below}")
            new_state.add(f"clear {below}")
            new_state.remove("arm-empty")
            new_state.remove(f"clear {block}")
            new_state.add(f"holding {block}")
            successors.append(new_state)

    def stack(block, target):
        if is_holding(block) and is_clear(target):
            new_state = state.copy()
            new_state.add(f"on {block} {target}")
            new_state.remove(f"clear {target}")
            new_state.add(f"clear {block}")
            new_state.add("arm-empty")
            new_state.remove(f"holding {block}")
            successors.append(new_state)

    # Try each action with each block
    blocks = {predicate.split()[-1] for predicate in state if "block" in predicate.split()[-1]}
    for block in blocks:
        pick_up(block)
        put_down(block)
        for other_block in blocks:
            if block != other_block:
                unstack(block, other_block)
                stack(block, other_block)

    return successors

# Example usage:
current_state = set(["clear A", "on-table A", "clear B", "on-table B", "arm-empty"])
successor_states(current_state)
\end{minted}

This updated version of the stack function now correctly adds the clear {block} predicate for the block being stacked, ensuring it's clear because there's nothing on top of it after the action. This update reflects the logic needed to maintain a consistent state representation after stacking operations.

\vspace{0.2cm}

{\em User}:
The blocks are not correctly identified from this state representation. Please fix
\vspace{0.2cm}

{\em ChatGPT}:
You're right; the approach for identifying blocks from the state representation in the previous code is flawed because it inaccurately captures block names when parsing predicates. I'll correct this by implementing a more reliable method to extract block names directly from the predicates involved in the state.

Here's the revised version of the successor\_states function that includes an improved way to determine the set of blocks based on the current state:

\begin{minted}{python}
def successor_states(state):
    successors = []

    # Helpers to check conditions and update states
    def is_clear(block):
        return f"clear {block}" in state

    def is_on_table(block):
        return f"on-table {block}" in state

    def is_holding(block):
        return f"holding {block}" in state

    def is_arm_empty():
        return "arm-empty" in state

    def find_block_on_top(block):
        for predicate in state:
            if predicate.startswith("on ") and predicate.endswith(block):
                return predicate.split()[1]
        return None

    # Actions
    def pick_up(block):
        if is_on_table(block) and is_clear(block) and is_arm_empty():
            new_state = state.copy()
            new_state.remove(f"on-table {block}")
            new_state.remove(f"clear {block}")
            new_state.remove("arm-empty")
            new_state.add(f"holding {block}")
            successors.append(new_state)

    def put_down(block):
        if is_holding(block):
            new_state = state.copy()
            new_state.add(f"on-table {block}")
            new_state.add(f"clear {block}")
            new_state.add("arm-empty")
            new_state.remove(f"holding {block}")
            successors.append(new_state)

    def unstack(block, below):
        if is_clear(block) and f"on {block} {below}" in state and is_arm_empty():
            new_state = state.copy()
            new_state.remove(f"on {block} {below}")
            new_state.add(f"clear {below}")
            new_state.remove("arm-empty")
            new_state.remove(f"clear {block}")
            new_state.add(f"holding {block}")
            successors.append(new_state)

    def stack(block, target):
        if is_holding(block) and is_clear(target):
            new_state = state.copy()
            new_state.add(f"on {block} {target}")
            new_state.remove(f"clear {target}")
            new_state.add(f"clear {block}")
            new_state.add("arm-empty")
            new_state.remove(f"holding {block}")
            successors.append(new_state)

    # Extract blocks from the state
    blocks = set()
    for predicate in state:
        parts = predicate.split()
        # Add block names (assuming the last word in each predicate is a block name or another block)
        if "table" not in parts[-1]:  # avoid adding 'table' as a block
            blocks.update(parts[1:])

    # Try each action with each block
    for block in blocks:
        # if "block" in block:  # Filter to ensure only block names are processed
        if True:
            pick_up(block)
            put_down(block)
            for other_block in blocks:
                if block != other_block:
                    unstack(block, other_block)
                    stack(block, other_block)

    return successors

# Example usage:
current_state = set(["clear A", "on-table A", "clear B", "on-table B", "arm-empty"])
successor_states(current_state)
\end{minted}

In this revised version, I have improved the block identification by iterating over all predicates and extracting block names from them directly, ensuring that only valid block names are considered for actions. The revised code assumes that block names can be identified as they don't contain the word "table", thereby excluding predicates like "on-table". This approach provides a more robust way to handle different naming conventions or additional predicates that might be added to the state.

\vspace{0.2cm}

{\em User}:
There is no need for checking if the string "block" is part of the block name, it might not be. Please update the code.
\vspace{0.2cm}

{\em ChatGPT}:
You're right; assuming that the term "block" will be a part of the block's name in predicates is restrictive and may not apply to all scenarios. Instead, we'll directly gather the block names by processing the appropriate parts of each predicate more accurately. Here’s an updated version of the successor\_states function that extracts block names based on the structure of predicates without assuming specific naming conventions:

\begin{minted}{python}
def successor_states(state):
    successors = []

    # Helpers to check conditions and update states
    def is_clear(block):
        return f"clear {block}" in state

    def is_on_table(block):
        return f"on-table {block}" in state

    def is_holding(block):
        return f"holding {block}" in state

    def is_arm_empty():
        return "arm-empty" in state

    def find_block_on_top(block):
        for predicate in state:
            if predicate.startswith("on ") and predicate.endswith(" " + block):
                return predicate.split()[1]
        return None

    # Actions
    def pick_up(block):
        if is_on_table(block) and is_clear(block) and is_arm_empty():
            new_state = state.copy()
            new_state.remove(f"on-table {block}")
            new_state.remove(f"clear {block}")
            new_state.remove("arm-empty")
            new_state.add(f"holding {block}")
            successors.append(new_state)

    def put_down(block):
        if is_holding(block):
            new_state = state.copy()
            new_state.add(f"on-table {block}")
            new_state.add(f"clear {block}")
            new_state.add("arm-empty")
            new_state.remove(f"holding {block}")
            successors.append(new_state)

    def unstack(block, below):
        if is_clear(block) and f"on {block} {below}" in state and is_arm_empty():
            new_state = state.copy()
            new_state.remove(f"on {block} {below}")
            new_state.add(f"clear {below}")
            new_state.remove("arm-empty")
            new_state.remove(f"clear {block}")
            new_state.add(f"holding {block}")
            successors.append(new_state)

    def stack(block, target):
        if is_holding(block) and is_clear(target):
            new_state = state.copy()
            new_state.add(f"on {block} {target}")
            new_state.remove(f"clear {target}")
            new_state.add(f"clear {block}")
            new_state.add("arm-empty")
            new_state.remove(f"holding {block}")
            successors.append(new_state)

    # Extract blocks from the state
    blocks = set()
    for predicate in state:
        parts = predicate.split()
        if parts[0] in ["on", "clear", "holding", "on-table"]:
            # Assuming the block names are always the last element in predicates that involve block names
            blocks.add(parts[-1])
            if parts[0] == "on":  # also add the block beneath if predicate starts with "on"
                blocks.add(parts[1])

    # Try each action with each block
    for block in blocks:
        pick_up(block)
        put_down(block)
        for other_block in blocks:
            if block != other_block:
                unstack(block, other_block)
                stack(block, other_block)

    return successors
\end{minted}

This version extracts block names from predicates based on their position in the string, which is typically the last part for predicates like "clear", "on-table", and "holding". For predicates starting with "on", both blocks involved are considered. This method ensures that all blocks are appropriately identified and handled without presuming the presence of specific substrings like "block" in the block names.

\subsubsection{Experiment Run 5.}
{\em User}:
I am playing with a set of blocks where I need to arrange the blocks into stacks. Here are the actions I can do

   Pick up a block
   Unstack a block from on top of another block
   Put down a block
   Stack a block on top of another block

   I have the following restrictions on my actions:
   I can only pick up or unstack one block at a time.
   I can only pick up or unstack a block if my hand is empty.
   I can only pick up a block if the block is on the table and the block is clear. A block is clear if the block has no other blocks on top of it and if the block is not picked up.
   I can only unstack a block from on top of another block if the block I am unstacking was really on top of the other block.
   I can only unstack a block from on top of another block if the block I am unstacking is clear.
   Once I pick up or unstack a block, I am holding the block.
   I can only put down a block that I am holding.
   I can only stack a block on top of another block if I am holding the block being stacked.
   I can only stack a block on top of another block if the block onto which I am stacking the block is clear.
   Once I put down or stack a block, my hand becomes empty.

If a state is defined by a list of following predicates "clear block", "on-table block", "arm-empty", "holding block", and "on block1 block2", how are successor states defined? Provide only the python code that returns a list of successor states for a state. 
\vspace{0.2cm}

{\em ChatGPT}:
To generate successor states based on the actions and restrictions you've described, you'll want a function that checks the current state (represented by a set of predicates), applies possible actions based on the rules, and then produces the new states. Here is a Python function that accomplishes this:

\begin{minted}{python}
def successor_states(state):
    # Break down the state into predicates
    clear = {b for b in state if b.startswith("clear ")}
    on_table = {b for b in state if b.startswith("on-table ")}
    arm_empty = "arm-empty" in state
    holding = {b[8:] for b in state if b.startswith("holding ")}
    on = {(b[3:], b2) for b in state for b2 in state if b.startswith("on ") and b.endswith(b2)}
    successors = []

    if arm_empty:
        # Pick up a block from the table
        for b in clear:
            block = b.split()[1]
            if f"on-table {block}" in state:
                new_state = state.copy()
                new_state.remove("arm-empty")
                new_state.remove(f"clear {block}")
                new_state.remove(f"on-table {block}")
                new_state.add(f"holding {block}")
                successors.append(new_state)

        # Unstack a block from another
        for (block1, block2) in on:
            if f"clear {block1}" in state:
                new_state = state.copy()
                new_state.remove("arm-empty")
                new_state.remove(f"clear {block1}")
                new_state.remove(f"on {block1} {block2}")
                new_state.add(f"clear {block2}")
                new_state.add(f"holding {block1}")
                successors.append(new_state)

    else:
        # Put down a block
        for block in holding:
            new_state = state.copy()
            new_state.add("arm-empty")
            new_state.add(f"clear {block}")
            new_state.add(f"on-table {block}")
            new_state.remove(f"holding {block}")
            successors.append(new_state)

        # Stack a block on top of another clear block
        for block in holding:
            for clear_block in clear:
                if clear_block.split()[1] != block:
                    new_state = state.copy()
                    new_state.add("arm-empty")
                    new_state.remove(f"clear {clear_block.split()[1]}")
                    new_state.remove(f"holding {block}")
                    new_state.add(f"on {block} {clear_block.split()[1]}")
                    new_state.add(f"clear {block}")
                    successors.append(new_state)

    return successors

current_state = set(["clear A", "on-table A", "clear B", "on-table B", "arm-empty"])
print(successor_states(current_state))

import parse_problem_blocks
def blocks_state_dict_to_list(state):
    # {'clear': ['b'], 'on-table': ['d'], 'arm-empty': True, 'holding': None, 'on': [('a', 'c'), ('b', 'a'), ('c', 'd')]}
    # {: ['d'], 'on': [('a', 'c'), ('b', 'a'), ('c', 'd')]}
    new_state = []
    if "arm-empty" in state and state["arm-empty"]:
        new_state.append("arm-empty")
    if 'holding' in state and state['holding'] is not None:
        new_state.append(f"holding {state['holding']}")

    for b in state['clear']:
        new_state.append(f"clear {b}")

    for b in state['on-table']:
        new_state.append(f"on-table {b}")

    for top, bottom in state['on']:
        new_state.append(f"on {top} {bottom}")

    return new_state


def blocks_state_dict_to_set(state):
    return set(blocks_state_dict_to_list(state))



init, goal = parse_problem_blocks.parse_pddl_blocks_from_file("blocks_instances/instance-0.pddl")
init = blocks_state_dict_to_set(init)
goal = blocks_state_dict_to_set(goal)
print("Init")
print(init)
print("Goal")
print(goal)
print("------------------")
curr = init.copy()
for i in range(10):
    print(curr)
    succ = successor_states(curr)
    print(succ)
    print("------------------")
    if len(succ) > 0:
        import random

        curr = random.choice(succ)
\end{minted}

This function assumes that the state parameter is a set of strings representing the predicates. The actions available depend on whether the robot's arm is empty or holding a block, and whether certain blocks are clear and available on the table or can be unstacked from another block. This function then modifies the state according to the rules of each action and adds each resulting new state to the list of successor states.

\vspace{0.2cm}

{\em User}:
Provide only the python code that tests whether a state is a goal state. Assume that the goal configuration is provided in the same format.
\vspace{0.2cm}

{\em ChatGPT}:
To test whether a given state matches a goal configuration, you can use a Python function that simply checks if all predicates specified in the goal are present in the current state. Here's a concise function to perform this check:

\begin{minted}{python}
def is_goal_state(state, goal):
    return goal.issubset(state)
\end{minted}

This function receives two sets: state and goal, where each is a set of strings representing the predicates. The function returns True if all predicates in the goal are contained within the current state, signifying that the state meets the goal configuration, and False otherwise. This approach is straightforward and leverages Python's set operations for efficiency.

\vspace{0.2cm}

{\em User}:
The successor\_states function does not correctly parse the state into predicates. Please rewrite. 
\vspace{0.2cm}

{\em ChatGPT}:
Certainly! It seems there was a misunderstanding in how predicates like "on" were being parsed. I'll adjust the parsing logic in the successor\_states function to correctly handle the predicates and relations between blocks. Here's the corrected version of the function:

\begin{minted}{python}
def successor_states(state):
    # Parse the current state
    clear_blocks = {b.split()[1] for b in state if b.startswith("clear")}
    on_table_blocks = {b.split()[1] for b in state if b.startswith("on-table")}
    arm_empty = "arm-empty" in state
    holding_blocks = {b.split()[1] for b in state if b.startswith("holding")}
    on_top_of = {(b.split()[1], b.split()[2]) for b in state if b.startswith("on ")}

    successors = []

    if arm_empty:
        # Pick up a block from the table
        for block in clear_blocks.intersection(on_table_blocks):
            new_state = state.copy()
            new_state.remove("arm-empty")
            new_state.remove(f"clear {block}")
            new_state.remove(f"on-table {block}")
            new_state.add(f"holding {block}")
            successors.append(new_state)

        # Unstack a block from another
        for (top, bottom) in on_top_of:
            if f"clear {top}" in state:
                new_state = state.copy()
                new_state.remove("arm-empty")
                new_state.remove(f"clear {top}")
                new_state.remove(f"on {top} {bottom}")
                new_state.add(f"clear {bottom}")
                new_state.add(f"holding {top}")
                successors.append(new_state)

    else:
        # Put down a block
        for block in holding_blocks:
            new_state = state.copy()
            new_state.add("arm-empty")
            new_state.add(f"clear {block}")
            new_state.add(f"on-table {block}")
            new_state.remove(f"holding {block}")
            successors.append(new_state)

        # Stack a block on top of another clear block
        for block in holding_blocks:
            for target in clear_blocks:
                if block != target:  # Cannot stack a block on itself
                    new_state = state.copy()
                    new_state.add("arm-empty")
                    new_state.remove(f"clear {target}")
                    new_state.remove(f"holding {block}")
                    new_state.add(f"on {block} {target}")
                    new_state.add(f"clear {block}")
                    successors.append(new_state)

    return successors
\end{minted}

In this revised function:

I ensure proper parsing of each predicate to correctly identify which blocks are clear, which are on the table, which one is being held, and the relationships between blocks (on top of).
Operations like picking up a block from the table and unstacking are allowed only when the arm is empty, and the corresponding block is clear and either on the table or on top of another block.
The stacking and putting down operations are processed when holding a block, respecting the restrictions on clear targets and table placements.
This code is more structured and should correctly handle the transition between states according to the provided rules and predicates.

\vspace{0.2cm}

{\em User}:
When getting the on\_top blocks from the state, we must differentiate the predicate "on x y" from "on-table x". They both start with "on". Please fix.
\vspace{0.2cm}

{\em ChatGPT}:
You're right; the initial parsing of the predicates didn't differentiate between "on-table" and "on" predicates correctly. I'll adjust the function to parse these predicates appropriately. Here's the updated successor\_states function:

\begin{minted}{python}
def successor_states(state):
    # Parse the current state
    clear_blocks = {b.split()[1] for b in state if b.startswith("clear")}
    on_table_blocks = {b.split()[1] for b in state if b.startswith("on-table")}
    arm_empty = "arm-empty" in state
    holding_blocks = {b.split()[1] for b in state if b.startswith("holding")}
    on_top_of = {(b.split()[1], b.split()[2]) for b in state if b.startswith("on ") and len(b.split()) == 3}

    successors = []

    if arm_empty:
        # Pick up a block from the table
        for block in clear_blocks.intersection(on_table_blocks):
            new_state = state.copy()
            new_state.remove("arm-empty")
            new_state.remove(f"clear {block}")
            new_state.remove(f"on-table {block}")
            new_state.add(f"holding {block}")
            successors.append(new_state)

        # Unstack a block from another
        for (top, bottom) in on_top_of:
            if f"clear {top}" in state:
                new_state = state.copy()
                new_state.remove("arm-empty")
                new_state.remove(f"clear {top}")
                new_state.remove(f"on {top} {bottom}")
                new_state.add(f"clear {bottom}")
                new_state.add(f"holding {top}")
                successors.append(new_state)

    else:
        # Put down a block
        for block in holding_blocks:
            new_state = state.copy()
            new_state.add("arm-empty")
            new_state.add(f"clear {block}")
            new_state.add(f"on-table {block}")
            new_state.remove(f"holding {block}")
            successors.append(new_state)

        # Stack a block on top of another clear block
        for block in holding_blocks:
            for target in clear_blocks:
                if block != target:  # Cannot stack a block on itself
                    new_state = state.copy()
                    new_state.add("arm-empty")
                    new_state.remove(f"clear {target}")
                    new_state.remove(f"holding {block}")
                    new_state.add(f"on {block} {target}")
                    new_state.add(f"clear {block}")
                    successors.append(new_state)

    return successors
\end{minted}

In this revised function:

The on\_top\_of set now ensures that only predicates starting with "on" and having exactly three components (which corresponds to "on X Y") are included.
This ensures accurate recognition and handling of the predicates representing blocks stacked on each other while avoiding confusion with "on-table" predicates.
The operations within the function continue to account for blocks being picked up, placed, or stacked according to the rules.


\clearpage
\subsection{PrOntoQA}

The prompt for the successor function is as follows.\\

{\em 
Logical deduction is performed by applying rules of the form if X then Y to our current knowledge about a world, extending our knowledge with each rule application. A rule if X then Y is applicable if we know X. If a state is defined as our current knowledge of the world, how are successor states defined, given a collection of rules? Provide only the Python implementation of successor states function
}

The prompt for the goal test is as follows.\\
{\em 
Provide only the python code that tests, given a target fact, whether a state is a goal state.
}

\subsubsection{Experiment Run 1.}

{\em User}:
Logical deduction is performed by applying rules of the form if X then Y to our current knowledge about a world, extending our knowledge with each rule application. A rule if X then Y is applicable if we know X. If a state is defined as our current knowledge of the world, how are successor states defined, given a collection of rules? Provide only the Python implementation.
\vspace{0.2cm}

{\em ChatGPT}:
In the context of logical deduction using a set of rules, a state can be considered as a set representing our current knowledge about the world. Each element in this set is a fact or a truth that we know about the world. Successor states are defined by applying applicable rules to the current state to extend our knowledge.

Here is a Python implementation to model this behavior. This implementation defines a function that takes the current state and a collection of rules, and produces the set of all successor states by applying these rules:

\begin{minted}{python}
def apply_rule(state, rule):
    """Apply a rule to the state if applicable."""
    if rule[0] in state:
        return state | {rule[1]}
    return state

def generate_successor_states(state, rules):
    """Generate all successor states by applying rules to the current state."""
    new_states = set()
    for rule in rules:
        new_state = apply_rule(state, rule)
        if new_state != state:  # Only add new states
            new_states.add(frozenset(new_state))
    return new_states

# Example usage
current_state = {'rain'}
rules = [('rain', 'wet'), ('sun', 'dry'), ('wet', 'puddles')]

# Generate successor states
successor_states = generate_successor_states(current_state, rules)
for state in successor_states:
    print(state)
    print(generate_successor_states(state, rules))
\end{minted}

In this code:

apply\_rule checks if the antecedent of a rule (first element of the tuple) is in the current state. If it is, it adds the consequent (second element) to the state.
generate\_successor\_states iterates over all rules and applies them to the current state, collecting new states that result from applying each rule. Each new state is added only if it is different from the current state, avoiding duplicates.
This model assumes that the rules are simple implication relationships and that the state can be represented as a set of facts. The generated successor states are those that include at least one new fact derived from applying the rules.

\vspace{0.2cm}

{\em User}:
Provide only the python code that tests, given a target fact, whether a state is a goal state.
\vspace{0.2cm}

{\em ChatGPT}:
To test whether a given state is a goal state based on the presence of a target fact, you can define a simple function in Python. This function will check if the target fact is in the current state. Here is how you can implement this:

\begin{minted}{python}
def is_goal_state(state, target_fact):
    """Check if the state contains the target fact."""
    return target_fact in state

# Example usage
state = {'rain', 'wet', 'puddles'}
target_fact = 'wet'

# Test if the state is a goal state
print(is_goal_state(state, target_fact))  # Output: True
\end{minted}

This function is\_goal\_state takes two arguments: the state (a set of facts) and the target\_fact (a fact that you want to check for). It returns True if the target fact is present in the state, indicating that this state is a goal state with respect to the target fact.

\subsubsection{Experiment Run 2.}

{\em User}:
Logical deduction is performed by applying rules of the form if X then Y to our current knowledge about a world, extending our knowledge with each rule application. A rule if X then Y is applicable if we know X. If a state is defined as our current knowledge of the world, how are successor states defined, given a collection of rules? Provide only the Python implementation of successor states function.
\vspace{0.2cm}

{\em ChatGPT}:
To implement the function for generating successor states using a logical deduction approach, you'll need a way to represent the current state of knowledge and the rules. Here's how you might do it in Python:

State Representation: Represent the state as a set of facts. Each fact can be a simple string or a more complex structure, depending on your application.
Rule Representation: Represent rules as tuples (X, Y), where X is a condition (or set of conditions) and Y is the conclusion. X can be a single fact or a set of facts.
Successor Function: The function checks each rule to see if its conditions are met in the current state. If so, it adds the rule's conclusion to the state, forming a new successor state.
Below is the Python implementation for generating successor states:

\begin{minted}{python}
def successor_states(current_state, rules):
    """
    Generate all successor states from the current state by applying applicable rules.
    
    Args:
    current_state (set): A set of facts known in the current state.
    rules (list of tuples): A list where each tuple represents a rule in the form (conditions, conclusion),
                            where `conditions` can be a single fact or a set of facts.
    
    Returns:
    list of sets: A list of all possible successor states.
    """
    new_states = []
    for conditions, conclusion in rules:
        # Check if the rule's conditions are met in the current state
        if isinstance(conditions, set):
            if conditions.issubset(current_state):
                # Apply the rule: create a new state with the conclusion added
                new_state = current_state.union({conclusion})
                new_states.append(new_state)
        else:
            if conditions in current_state:
                new_state = current_state.union({conclusion})
                new_states.append(new_state)

    return new_states
\end{minted}

This function checks each rule to see if the conditions are met within the current state. If they are, it creates a new state that includes the conclusion of the rule. This implementation also supports both single conditions and sets of conditions. Adjustments might be needed depending on the specific requirements and the complexity of the conditions and conclusions.

\vspace{0.2cm}

{\em User}:
Provide only the python code that tests, given a target fact, whether a state is a goal state.
\vspace{0.2cm}

{\em ChatGPT}:
Here's a Python function that tests whether a given state contains a target fact, effectively checking if it is a goal state:

\begin{minted}{python}
def is_goal_state(state, target_fact):
    """
    Check if the state contains the target fact, indicating it is a goal state.
    
    Args:
    state (set): The current state represented as a set of facts.
    target_fact (str): The fact that defines the goal state.
    
    Returns:
    bool: True if the target fact is in the state, False otherwise.
    """
    return target_fact in state
\end{minted}

This function simply checks if the target\_fact is a member of the state set, returning True if the state meets the goal criteria (i.e., contains the target fact) and False otherwise.

\subsubsection{Experiment Run 3.}

{\em User}:
Logical deduction is performed by applying rules of the form if X then Y to our current knowledge about a world, extending our knowledge with each rule application. A rule if X then Y is applicable if we know X. If a state is defined as our current knowledge of the world, how are successor states defined, given a collection of rules? Provide only the Python implementation of successor states function.
\vspace{0.2cm}

{\em ChatGPT}:
To provide a Python implementation for defining successor states based on a given collection of rules and a current state of knowledge, we need a function that takes the current state (a set of known facts) and the rules (formulated as "if X then Y") and produces all possible successor states by applying these rules.

Here's a straightforward Python function to achieve this:

\begin{minted}{python}
def successor_states(current_state, rules):
    """
    Generates all possible successor states by applying rules to the current state.

    :param current_state: A set of strings representing known facts.
    :param rules: A list of tuples (X, Y) representing the rules 'if X then Y'.
    :return: A set of sets, each representing a possible successor state.
    """
    new_states = set()  # To store unique successor states

    # Iterate through each rule and apply it if possible
    for x, y in rules:
        # Check if the condition of the rule is satisfied in the current state
        if x in current_state:
            # Create a new state by adding the consequence of the rule to the current state
            new_state = current_state.copy()
            new_state.add(y)
            new_states.add(frozenset(new_state))  # Add as frozenset for immutability

    return new_states

# Example usage:
current_state = {'X'}
rules = [('X', 'Y'), ('Y', 'Z')]
print(successor_states(current_state, rules))
\end{minted}

This function works by iterating through each rule and checking if the antecedent (condition) of the rule is already a known fact in the current\_state. If it is, it creates a new state by adding the consequent (result) of the rule to a copy of the current state. Each new state is added to a set of successor states, ensuring that only unique states are stored. Each state is stored as a frozenset to ensure immutability and to allow sets of sets.

\vspace{0.2cm}

{\em User}:
Provide only the python code that tests, given a target fact, whether a state is a goal state.
\vspace{0.2cm}

{\em ChatGPT}:
Here's a Python function to check whether a given state includes a specific target fact, thus determining if it's a goal state:

\begin{minted}{python}
def is_goal_state(state, target_fact):
    """
    Checks if the target fact is in the given state.

    :param state: A set of strings representing the current state.
    :param target_fact: A string representing the target fact to be achieved.
    :return: True if the target fact is in the state, False otherwise.
    """
    return target_fact in state

# Example usage:
state = {'X', 'Y', 'Z'}
target_fact = 'Y'
print(is_goal_state(state, target_fact))  # Output: True
\end{minted}

\vspace{0.2cm}

{\em User}:
In successor\_states, I get AttributeError: 'frozenset' object has no attribute 'add'
\vspace{0.2cm}

{\em ChatGPT}:
The error arises because I incorrectly used frozenset for the new\_state, which is immutable and doesn't support the add method. We should first modify the mutable set before converting it to a frozenset when adding it to new\_states. Let's correct the implementation:

\begin{minted}{python}
def successor_states(current_state, rules):
    """
    Generates all possible successor states by applying rules to the current state.

    :param current_state: A set of strings representing known facts.
    :param rules: A list of tuples (X, Y) representing the rules 'if X then Y'.
    :return: A set of frozensets, each representing a possible successor state.
    """
    new_states = set()  # To store unique successor states

    # Iterate through each rule and apply it if possible
    for x, y in rules:
        # Check if the condition of the rule is satisfied in the current state
        if x in current_state:
            # Create a new state by adding the consequence of the rule to the current state
            new_state = set(current_state)  # Make a mutable copy of the current state
            new_state.add(y)  # Add the new fact
            new_states.add(frozenset(new_state))  # Store as frozenset for immutability

    return new_states

# Example usage:
current_state = {'X'}
rules = [('X', 'Y'), ('Y', 'Z')]
print(successor_states(current_state, rules))
\end{minted}

This revised function ensures that the current\_state is copied into a mutable set before adding new elements. After the modifications are done, it is then converted into a frozenset for storage in the new\_states set. This approach maintains the immutability of states within new\_states and avoids errors related to modifying a frozenset.
\subsubsection{Experiment Run 4.}

{\em User}:
Logical deduction is performed by applying rules of the form if X then Y to our current knowledge about a world, extending our knowledge with each rule application. A rule if X then Y is applicable if we know X. If a state is defined as our current knowledge of the world, how are successor states defined, given a collection of rules? Provide only the Python implementation of successor states function
\vspace{0.2cm}

{\em ChatGPT}:
To implement a Python function for generating successor states from a given state using a set of logical deduction rules, we'll assume that the state is represented as a set of facts (truths known about the world), and the rules are represented as tuples (X, Y), meaning "if X then Y". A rule is applicable if all elements of X are present in the current state. When a rule is applied, Y is added to the state to create a successor state. Here's how this can be implemented:

\begin{minted}{python}
def successor_states(current_state, rules):
    """
    Generate all successor states from the current state using the given rules.
    
    :param current_state: A set of facts known about the world.
    :param rules: A list of tuples (X, Y), where each X is a tuple representing conditions and Y is a single fact.
    :return: A set of new states, each differing from the current by the addition of a new fact derived from a rule.
    """
    new_states = set()

    # Check each rule to see if it can be applied
    for conditions, result in rules:
        if all(condition in current_state for condition in conditions):
            # If all conditions are met, create a new state by adding the result to the current state
            new_state = current_state.copy()
            new_state.add(result)
            new_states.add(frozenset(new_state))  # Use frozenset to ensure the state can be added to a set

    return new_states

# Example usage
current_state = {'rainy', 'umbrella_open'}
rules = [(({ 'rainy' }, 'wet_ground'), ({ 'umbrella_open', 'rainy' }, 'dry_under_umbrella'))]

# Convert rule format to match the function expectation
formatted_rules = [((condition,), result) for condition, result in rules]

new_states = successor_states(current_state, formatted_rules)
print("New States:", [set(state) for state in new_states])
\end{minted}

In this example:

current\_state contains facts 'rainy' and 'umbrella\_open'.
rules specify that if it is 'rainy', then the 'wet\_ground' should be true, and if both 'umbrella\_open' and 'rainy' are true, then 'dry\_under\_umbrella' should be true.
The function successor\_states checks each rule, determines if the conditions are met, and generates new states by adding the results of applicable rules. Each new state is a superset of the current state with one additional fact.

\vspace{0.2cm}

{\em User}:
Provide only the python code that tests, given a target fact, whether a state is a goal state.
\vspace{0.2cm}

{\em ChatGPT}:
Here's the Python code that tests whether a given state contains a target fact, thereby determining if it is a goal state:

\begin{minted}{python}
def is_goal_state(state, target_fact):
    """
    Determine if the given state is a goal state based on the presence of the target fact.
    
    :param state: A set of facts known about the world.
    :param target_fact: The fact that if present in the state, identifies it as a goal state.
    :return: True if the target fact is in the state, False otherwise.
    """
    return target_fact in state

# Example usage
state = {'rainy', 'wet_ground', 'umbrella_open'}
target_fact = 'wet_ground'
result = is_goal_state(state, target_fact)
print("Is goal state:", result)
\end{minted}

This function simply checks if the target\_fact is a part of the current state. If it is, the function returns True, indicating that the state is a goal state; otherwise, it returns False.

\vspace{0.2cm}

{\em User}:
In successor\_states function, assume that the rules are just pairs of facts, so a single fact condition. Also, you cannot add to a frozenset
\vspace{0.2cm}

{\em ChatGPT}:
Given the revised assumptions where each rule is just a pair of facts and each condition is represented by a single fact, and considering that we cannot add to a frozenset, we can refactor the successor\_states function. The function will now operate on a copy of the current state converted to a mutable set, apply the changes, and then convert back to a frozenset for immutability and use in a set of states. Here's the updated implementation:

\begin{minted}{python}
def successor_states(current_state, rules):
    """
    Generate all successor states from the current state using the given rules, where each rule is a pair (X, Y),
    meaning if X then Y. Each X is a single fact condition.
    
    :param current_state: A frozenset representing the current state of known facts.
    :param rules: A list of tuples (X, Y), where X is a fact and Y is a fact that follows if X is true.
    :return: A set of new states, each differing from the current by the addition of a new fact derived from a rule.
    """
    new_states = set()
    mutable_current_state = set(current_state)  # Convert frozenset to a mutable set

    # Check each rule to see if it can be applied
    for condition, result in rules:
        if condition in mutable_current_state:
            # If condition is met, create a new state by adding the result to the current state
            new_state = mutable_current_state.copy()
            new_state.add(result)
            new_states.add(frozenset(new_state))  # Convert back to frozenset for immutability

    return new_states

# Example usage
current_state = frozenset({'rainy', 'umbrella_open'})
rules = [('rainy', 'wet_ground'), ('umbrella_open', 'dry_under_umbrella')]

new_states = successor_states(current_state, rules)
print("New States:", [set(state) for state in new_states])
\end{minted}

This code:

Converts the immutable frozenset of the current state to a mutable set to allow modifications.
Iterates over each rule, checking if the condition is met in the current state.
If the condition is met, it creates a new state by copying the current state, adding the result of the rule, and then converting it back to a frozenset for use in a set (to ensure immutability and unique state management).
Returns a set of all new states generated this way.

\subsubsection{Experiment Run 5.}
{\em User}:
Logical deduction is performed by applying rules of the form if X then Y to our current knowledge about a world, extending our knowledge with each rule application. A rule if X then Y is applicable if we know X. If a state is defined as our current knowledge of the world, how are successor states defined, given a collection of rules? Provide only the Python implementation of successor states function
\vspace{0.2cm}

{\em ChatGPT}:
In a logical deduction system, we can model successor states using a function that takes the current state of knowledge (a set of facts) and a collection of rules. Each rule can be represented as a tuple (X, Y) where if X (a precondition) is satisfied by the current state, then Y (a consequence) can be added to the state.

Here's a Python function that computes the successor states by applying all applicable rules to the current state:

\begin{minted}{python}
def apply_rules(state, rules):
    """
    Apply rules to the current state to generate successor states.
    
    :param state: A set representing the current state (known facts)
    :param rules: A list of tuples, each tuple is a rule (X, Y) where
                  if X is in the state, Y is added to the state
    :return: A new state formed by applying all applicable rules
    """
    new_state = set(state)  # Start with a copy of the current state
    changes = True          # To check if any new fact was added
    
    while changes:
        changes = False
        for precondition, consequence in rules:
            if precondition in new_state and consequence not in new_state:
                new_state.add(consequence)
                changes = True  # A new fact was added, continue the loop
    
    return new_state

# Example usage
current_state = {'X'}
rules = [('X', 'Y'), ('Y', 'Z')]
new_state = apply_rules(current_state, rules)
print(new_state)  # Output should be {'X', 'Y', 'Z'}
\end{minted}

This function works as follows:

It initializes the new state with the current state.
It iterates over the rules and checks if the precondition of any rule is met by the current state.
If a precondition is met and the consequence is not already in the state, the consequence is added.
This process repeats until no new facts can be added, ensuring all applicable rules have been exhausted.
The function ensures that all direct and indirect consequences of the initial facts are considered, making it useful for computing successor states in logical deduction systems.

\vspace{0.2cm}

{\em User}:
Provide only the python code that tests, given a target fact, whether a state is a goal state.
\vspace{0.2cm}

{\em ChatGPT}:
Certainly! Below is a Python function that tests whether a given state contains a target fact, which would indicate it's a goal state:

\begin{minted}{python}
def is_goal_state(state, target_fact):
    """
    Check if the state is a goal state by determining if it contains the target fact.

    :param state: A set representing the current state (known facts)
    :param target_fact: The fact that defines the goal state
    :return: True if the target_fact is in the state, False otherwise
    """
    return target_fact in state

# Example usage
state = {'X', 'Y', 'Z'}
target_fact = 'Y'
result = is_goal_state(state, target_fact)
print(result)  # Output should be True

target_fact = 'A'
result = is_goal_state(state, target_fact)
print(result)  # Output should be False
\end{minted}

This function takes a set state containing all known facts and checks if the target\_fact is among them, thus determining if the state is a goal state.

The implementation of apply\_rules applies all possible rules iteratively. That is not what I asked for, I asked for a successor states function, returning all immediate successor states
\vspace{0.2cm}

{\em ChatGPT}:
Thank you for the clarification. To implement a function that generates all immediate successor states, we need a different approach. Instead of applying all rules iteratively until no more can be applied, we'll generate a new state for each single rule application, thereby obtaining all the states that are one rule application away from the current state.

Here's the Python implementation for the function that generates all immediate successor states:

\begin{minted}{python}
def successor_states(state, rules):
    """
    Generate all immediate successor states by applying each applicable rule once.

    :param state: A set representing the current state (known facts)
    :param rules: A list of tuples, each tuple is a rule (X, Y) where
                  if X is in the state and Y is not, Y is added to the state
    :return: A list of new states, each created by applying one rule to the current state
    """
    successors = []
    
    for precondition, consequence in rules:
        if precondition in state and consequence not in state:
            new_state = set(state)
            new_state.add(consequence)
            successors.append(new_state)
    
    return successors

# Example usage
current_state = {'X'}
rules = [('X', 'Y'), ('Y', 'Z'), ('X', 'A')]
all_successors = successor_states(current_state, rules)

for idx, state in enumerate(all_successors):
    print(f"Successor {idx + 1}: {state}")
\end{minted}

This function iterates through each rule and checks if the precondition is met and the consequence is not already in the state. If so, it creates a new state with the consequence added and adds this new state to the list of successors. Each state in the list represents a distinct state reached by applying one rule from the initial state.



\end{document}